\documentclass[journal]{IEEEtran}
\usepackage{tikz}
\usetikzlibrary{bayesnet}
\usepackage{epsfig}
\usepackage{graphicx}
\usepackage[cmex10]{amsmath}
\usepackage{amssymb}
\usepackage{algorithm}
\usepackage{algpseudocode}
\usepackage{ctable}
\usepackage{multirow}
\usepackage{epstopdf}
\usepackage{color, soul}
\usepackage{enumitem}

\usepackage{array}
\usepackage{hyperref}
\let\oldemptyset\emptyset
\let\emptyset\varnothing
\setlist[itemize]{leftmargin=*}

\usepackage[para,online,flushleft]{threeparttable}

\ifCLASSINFOpdf
\else
\fi

\begin{document}
%

%
%

\date{}

\title{Multi-Instance Dynamic Ordinal Random Fields for Weakly-supervised Facial Behavior Analysis}

\author{Adria~Ruiz$^*$,
        Ognjen (Oggi)~Rudovic$^\dagger$,
        Xavier~Binefa$^*$ and
        Maja~Pantic$^\diamond$
\IEEEcompsocitemizethanks{\IEEEcompsocthanksitem $^*$ A. Ruiz and X. Binefa are at the Cognitive Media Technology Group, Pompeu Fabra University, Spain (\href{mailto:adria.ruiz@upf.edu}{adria.ruiz@upf.edu},\href{mailto:xavier.binefa@upf.edu}{xavier.binefa@upf.edu}) }
\IEEEcompsocitemizethanks{\IEEEcompsocthanksitem $^\dagger$ O. Rudovic is at MIT Media Lab, Cambridge, USA (\href{mailto:orudovic@media.mit.edu}{orudovic@mit.edu}) }
\IEEEcompsocitemizethanks{\IEEEcompsocthanksitem $^\diamond$ M. Pantic is at Dept. of Computing, Imperial College London, UK. She is also with the Faculty of Electrical Engineering, Mathematics and Computer Science, University of Twente, The Netherlands (\href{mailto:m.pantic@imperial.ac.uk}{m.pantic@imperial.ac.uk}) }
}
\maketitle

\IEEEtitleabstractindextext{%

\begin{abstract}
We propose a Multi-Instance-Learning (MIL) approach for weakly-supervised learning problems, where a training set is formed by bags (sets of feature vectors or instances) and only labels at bag-level are provided. Specifically, we consider the Multi-Instance Dynamic-Ordinal-Regression (MI-DOR) setting, where the instance labels are naturally represented as ordinal variables and bags are structured as temporal sequences. To this end, we propose Multi-Instance Dynamic Ordinal Random Fields (MI-DORF). In this framework, we treat instance-labels as temporally-dependent latent variables in an Undirected Graphical Model. Different MIL assumptions are modelled via newly introduced high-order potentials relating bag and instance-labels within the energy function of the model. We also extend our framework to address the Partially-Observed MI-DOR problems, where a subset of instance labels are available during training. We show on the tasks of weakly-supervised facial behavior analysis, Facial Action Unit (DISFA dataset) and Pain (UNBC dataset) Intensity estimation, that the proposed framework outperforms  alternative learning approaches. Furthermore, we show that MI-DORF can be employed to reduce the data annotation efforts in this context by large-scale. 
\end{abstract}

\begin{IEEEkeywords}
Mutiple Instance Learning, Undirected Graphical Models, Facial Behavior Analysis, Pain Intensity, Action Units
\end{IEEEkeywords}}


\IEEEdisplaynontitleabstractindextext

%
\IEEEpeerreviewmaketitle






\section{Introduction}
\label{sec:introduction}

Mutli-Instance-Learning (MIL) is a popular modelling framework for addressing different weakly-supervised problems \cite{babenko2011robust,wu2014milcut,ruiz2014regularized}. In traditional Single-Instance-Learning (SIL), the fully supervised setting is assumed with the goal to learn a model from a set of feature vectors (instances) each being annotated in terms of target label $y$. By contrast, in MIL, the weak supervision is assumed, thus, the training set is formed by bags (sets of instances), and only labels at bag-level are provided. In order to learn a model from this weak-information, MIL assumes that there exists an underlying relation between the label of a bag (e.g., video) and the labels of its constituent instances (e.g., image frames). For instance, in standard Multi-Instance-Classification (MIC) \cite{maron1998framework}, labels are considered binary variables $y \in \{-1,1\}$ and negative bags are assumed to contain only instances with an associated negative label. In contrast, positive bags must contain at least one positive instance. Another example of MIL assumption is related to the Multi-Instance-Regression (MIR) problem \cite{ray2001multiple}, where $y \in R$ is a real-valued variable and the maximum instance-label within the bag is assumed to be equal to $y$. Different from previous works, in this paper we focus on a novel MIL problem that we refer to as Multi-Instance Dynamic Ordinal Regression (MI-DOR). In this case, bags are structured as  dynamic sequences of instances with temporal dependencies. Moreover, instance labels are considered ordinal variables which can take values in a set of $L$ discrete categories satisfying the increasing monotonicity constraints  $\{ 0 \prec ... \prec l \prec L \}$. Our definition of MI-DOR is enough general to define different weak-relations between bag and instance-labels. {Specifically, we focus on two instances of this problem: Maximum and Relative MI-DOR. Similar to MIR, in the former, we assume that the maximum ordinal value within a sequence is equal to its bag (sequence) label. On the other hand, the latter assumes that the weak-label provides information about the evolution (increase, decrease or monotone) of the instance ordinal levels within the sequence. As we discuss below, these two have important applications in the context of Facial Behavior Analysis that we address in this paper.}

\subsection{Motivation: Weakly-Supervised Facial Behavior Analysis}
\label{sec:introduction_motiv}

\begin{figure}[t]
    \centering
    \includegraphics[width=0.4\textwidth]{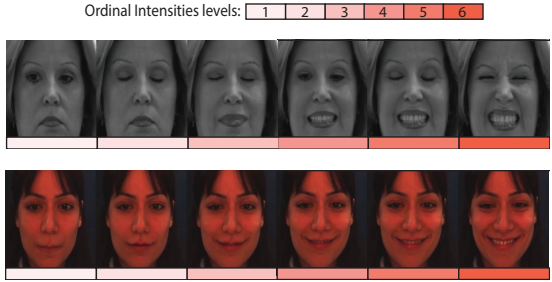}
    \caption{Illustration of the Pain and Action Unit intensity problems addressed in this work. Top: Sequence showing different pain levels (coded in an ordinal scale from 1 to 6). Bottom: Example of different intensities for Action Unit 12 (Lip-Corner Puller) also represented in an ordinal scale. }
    \label{fig:aupain_problem}
\end{figure}

Facial expressions provide information about human emotions, attitudes and mental states \cite{ekman1997face}. Their automatic analysis  has become a very active research field in Computer Vision in the last decade due to the large number of potential applications in different contexts such as medicine or entertainment. In this work, we focus on two relevant problems of automatic facial behavior analysis: Action Unit (AU) \cite{mavadati2013disfa} and Pain \cite{lucey2011painful} Intensity estimation. Both can be naturally posed as Dynamical Ordinal Regression problems, where the goal is to predict a value on an ordinal scale for each instant of a sequence. Specifically, in AU intensity estimation, the objective is to predict the activation level (on a six-point ordinal scale) of facial actions at each frame in a video. Similarly, in the Pain Intensity estimation task we aim to measure the intensity level of pain felt by a patient (see Fig.~\ref{fig:aupain_problem}).

One potential solution addressing this limitation could be to annotate larger training sets. However, this strategy is not feasible given the expense of the annotation process. In contrast, the explored solution in this work consists of using the weakly-supervised paradigm instead of the fully-supervised one. Weakly-supervised approaches aim to learn models using annotations which only provide partial information (weak-labels) about the task that needs to be solved. These weak-labels are much easier to obtain than those for fully-supervised learning, thus allowing us to use larger datasets minimizing the annotation effort. For example, in Pain Intensity estimation, it is much easier to obtain a label for the whole sequence in terms of the maximum pain intensity felt by the recorded subject (e.g. using patients self-reports or external observers). Similarly, annotating  Facial Action Unit intensities requires a huge effort by expert coders. In contrast, segmenting sequences according to the increasing or decreasing evolution of AU intensities (i.e, onset and appex segments) is less time-consuming. {These two scenarios motivates our interest in the Maximum and Relative MI-DOR problems previously introduced}. Models able to learn only from these weak information would allow to leverage larger training sets and thus potentially build more effective models for intensity estimation of different facial behaviours.
\label{sec:motivation}

\subsection{Contributions}
In this work, we propose the Multi-Instance Dynamic Ordinal Random Fields (MI-DORF) framework to address MI-DOR problems. To build our approach, we use the notion of  Hidden Conditional Ordinal Random Fields (HCORF) \cite{kim2010hidden}. Similar to HCORF, MI-DORF is an Undirected Graphical Model where observation labels are modelled as a linear-chain of ordinal latent variables. However, the energy function of MI-DORF is designed to explicitly incorporate the Multiple Instance relation between latent instance labels and observable sequence weak-labels. The main contributions of this work can be summarized as follows:
\begin{itemize}
\item To the best of our knowledge, no previous works have explored Multi-Instance Dynamic Ordinal Regression problems {(Sec. \ref{sec:MIDOR}). The proposed MI-DORF framework addresses these tasks by explicitly modelling the weak-relation between instances and sequence labels.}. Our framework is the first MIL approach that imposes ordinal constraints on the instance labels. The proposed method also incorporates dynamic information that is important when modeling temporal structure in instances within the bags (i.e., image sequences). While modeling dynamic information has been attempted in \cite{wu2015multi,liu2015video}, there are virtually no works that account for both ordinal and temporal data structures within existing MIL frameworks. 

\item {We  also introduce high-order potentials in the MI-DORF energy function in order to model weakly-supervised MIL assumptions. Following this strategy, we present two variants of this framework: MaxMI-DORF {(Sec.~\ref{sec:MMIDORF} )} and RelMI-DORF {(Sec. \ref{sec:RMIDORF} )}. A preliminary version of the particular MaxMI-DORF method was presented in our previous work. \cite{ruiz2016multi}. These two models are specially designed to address the Maximum and Relative MI-DOR problems, respectively. Given that the newly introduced MIL potentials of our models render the standard inference procedures for existing latent variable models (e.g., HCORF) infeasible, we derive a novel inference procedure. This procedure scales well with the data number and its computational complexity is similar to that of forward-backward algorithm \cite{barber2012bayesian}, typically employed in linear-chains models.}

\item We also propose the Partially-Observed extension of our MI-DORF model (Sec VI). This approach allows us to leverage  available instance labels in order to increase the level of supervision in our model. To this end, we generalize the learning and inference procedures of the MI-DORF models mentioned above, making them applicable to the partially-observed and still weakly-supervised learning tasks. We show that with a small portion of labeled instances, we can reach the performance of the fully supervised models for target tasks, thus, reducing the expensive (manual) data annotation efforts by large-scale.
\end{itemize}

We demonstrate the performance of the proposed methods on weakly-supervised Pain {(Sec. \ref{sec:au_exp} )} and Action Unit Intensity estimation using the benchmark datasets for target tasks. {(Sec. \ref{sec:pain_exp} )}. We show under various settings  the advantages of our method compared to alternative approaches.

\label{sec:contributions}


\section{Related Work}
\label{sec:related}

\subsection{Multiple-Instance Learning } 
Existing  MIL approaches usually follow the bag-based or instance-based paradigms \cite{amores2013multiple}. In the bag-based methods, a feature vector representation for each bag is first extracted. Then, these representations are used to train standard Single-Instance methods, used to estimate the bag labels. This representation is usually computed by using different types of similarity metrics between training instances. Examples following this paradigm include Multi-Instance Kernel \cite{gartner2002multi}, MILES \cite{chen2006miles} or MI-Graph \cite{zhou2009multi}. The main limitation of these approaches is that the learned models can only make predictions at the bag-level (e.g., a video) and are not able to estimate instance-labels (e.g., frame-level intensities). In contrast, instance-based methods directly learn a model which operates at the instance level. For this, MIL assumptions are incorporated by considering instance-labels as latent variables. Using this strategy,  traditional supervised models are adapted to incorporate MIL assumptions. Examples of methods following this approach include Multi-Instance Support Vector Machines \cite{andrews2002support} (MI-SVM), MILBoost \cite{zhang2005multiple}, MI Gaussian Processes \cite{kim2010gaussian} or MI Logistic Regression \cite{hsu2014augmented}. In this work, we follow the instance-based paradigm by treating instance-labels as ordinal latent states in a Latent-Dynamic Model. In particular, we follow a similar idea to that in the Multi-Instance Discriminative Markov Networks \cite{hajimirsadeghi2013multiple}, where the energy function of a Markov Network is designed to explicitly model weak-relations between bag and instance labels. However, in contrast to the works described above, the presented MI-DORF framework accounts for the ordinal structure in instance labels, while also accounting for their dynamics.

\subsection{Latent-Dynamic Models}
 Popular methods for sequence classification are Latent-Dynamic Models such as Hidden Conditional Random Fields (HCRFs) \cite{quattoni2007hidden} or Hidden-Markov-Models (HMMs) \cite{rabiner1986introduction}. These methods are variants of Dynamic Bayesian Networks (DBNs) where a set of latent states are used to model the conditional distribution of observations given the sequence label. In these approaches, dynamic information is modelled by incorporating probabilistic dependence between time-consecutive latent states. MI-DORF builds upon the HCORF framework \cite{kim2010hidden} which considers latent states as ordinal variables. However, HCORF follows the supervised paradigm, where the main goal is to predict sequence labels and latent variables are only used to increase the expressive power of the model. In contrast, the energy function of MI-DORF is defined to explicitly encode Multi-Instance relationships between bag and latent instance labels. Note also that more recent works (e.g., \cite{wu2015multi}, \cite{liu2015video}) extended HMMs/HCRFs, respectively, for Multi Instance Classification. The reported results in these works suggested that modeling dynamics in MIL can be beneficial when bag-instances exhibit temporal structure. However, these methods limit their consideration to the case where instance labels are binary and, therefore, are unable to solve MI-DOR problems. 
 
 As has been introduced in Sec. \ref{sec:contributions}, we also extend MI-DORF to the partially-observed setting, where labels for a small subset of instances are available during training. This scenario has been previously explored using Latent-dynamical models such as Conditional Random Fields \cite{li2009extracting} and their extensions (HCRF \cite{chang2009learning}). Although the instance labels are incorporated in these approaches, they can be considered suboptimal for MI-DOR, where sequence weak-labels need to be also taken into account according to the MIL assumptions. 
 
\subsection{Non-supervised facial behavior analysis}

{Research on automatic facial behavior analysis has mainly focused on the fully-supervised setting. In the specific problems of Action Unit and Pain Intensity Estimation, recent works have developed models based on HCORF \cite{rudovic2015context}, Metric Learning \cite{nicolle2016real}, Convolutional Neural Networks \cite{walecki2017_deep_copula} or Gaussian Processes \cite{eleftheriadis_accv} among others. However, as discussed in Sec \ref{sec:introduction}, supervised models are limited in this context because they involve a laborious data labelling.} 

{In order to reduce the annotation efforts, in this work we address these problems using weakly-supervised learning, which lies on the spectrum in between the unsupervised and fully supervised paradigms. In this context, previous works have explored non-supervised approaches for Facial Behavior Analysis.} For AU detection, Zhou et. al \cite{zhou2010unsupervised} proposed Aligned Cluster Analysis for the unsupervised segmentation and clustering of facial events in videos. Their experiments showed that the obtained clusters were coherent with AU manual annotations. We find another example in \cite{tax2010detection}, where Multiple Instance Classification was used to find key frames representing Action Unit activations in sequences. Different from these cited approaches which focus on binary detection, we address weakly-supervised Action Unit intensity estimation. To this end, the proposed MI-DORF model is able to learn from segments which are labelled according to the increasing or decreasing evolution of AU intensities (see Sec. \ref{sec:motivation}). A similar problem has been recently addressed by Zhao et al. \cite{Zhao_2016_CVPR}. Specifically, Ordinal Support Vector Ordinal Regression (OSVR) was used to estimate facial expression intensities using only onset and appex segments during training. However, OSVR presents some limitations in this context. Firstly, it models the instance (frame) labels as continuous variables which is a sub-optimal modelling of ordinal variables. Secondly, OSVR poses MI-DOR as a ranking problem causing  the scale of predicted values to not necessarily match with the ground-truth. In contrast, MI-DORF models instance labels as ordinal variables, thus allowing to better estimate labels scale by determining a priory the number of ordinal levels. Finally, OSVR is an static approach and temporal correlations are not modelled as in MI-DORF.

 In the context of weakly-supervised Pain Intensity estimation,  MIL approaches have been previously applied by considering that a weak-label is provided for a sequence (in terms of the maximum pain intensity felt by the patient). Then, a video is considered as a bag and image frames as instances. Sikka et al. \cite{sikka2013weakly} proposed to extract a Bag-of-Words representation from video segments and treat them as bag-instances. Then, MILBoosting \cite{zhang2005multiple} was applied to predict sequence-labels under the MIC assumption. Following the bag-based paradigm, \cite{ruiz2014regularized} developed the Regularized Multi-Concept MIL method capable of discovering different discriminative pain expressions within a sequence. More recently, \cite{wu2015multi} proposed  MI Hidden Markov Models, an adaptation of standard HMM to the MIL problem. The limitation of these approaches is that they focus on the binary detection problem (i.e, pain intensity levels are binarized) and thus, are unable to consider different intensity levels of pain. This is successfully attained by the proposed MI-DORF.

\begin{figure*}[ht]
    \centering
    \begin{tabular}{c c  c c}
        \resizebox{0.45\textwidth}{!}{
\begin{tikzpicture}
  \node[obs]                                              (x1) {$x_1$};
  \node[obs, right=1.6 of x1]                               (x2) {$x_2$};
  \node[obs, right=1.6 of x2]                               (x3) {$x_3$};
  \node[obs, right=1.6 of x3]                               (xt) {$x_t$};
  \node[obs, right=3.2 of xt]                               (xT) {$x_T$};

  \node[latent, above=1 of x1]                           (h1) {$h_1$};
  \node[latent, right=1.6 of h1]                           (h2) {$h_2$};
  \node[latent, right=1.6 of h2]                           (h3) {$h_3$};
  \node[latent, right=1.6 of h3]                           (ht) {$h_t$};
  \node[latent, right=3.2 of ht]                           (hT) {$h_T$};

  \node[obs, above=0.8 of h3]                          (y) {$y$};

  \factor[below=0.25 of h1] {} {right:$\Psi^{N}(\mathbf{x},h)$} {h1,x1} {} ; %
  \factor[below=0.25 of h2] {} {} {h2,x2} {} ; %
  \factor[below=0.25 of h3] {} {} {h3,x3} {} ; %
  \factor[below=0.25 of ht] {} {} {ht,xt} {} ; %
  \factor[below=0.25 of hT] {} {} {hT,xT} {} ; %

  \factor[right=0.6 of h1] {} {} {h1,h2} {} ; %
  \factor[right=0.6 of h2] {} {} {h2,h3} {} ; %
  \factor[right=0.6 of h3] {} {} {h3,ht} {} ; %
  \factor[right=1.3 of ht] {} {below=1:$\Psi^E(h_t,h_{t+1},y)$} {ht,hT} {} ; %
 
 \factor[below=0.3 of y] {} {above:\hspace{2cm}$\Psi^{M}(\mathbf{h},y)$} {h1,h2,h3,ht,hT,y} {} ; %

  \end{tikzpicture}
} & ~ & ~ \resizebox{0.45\textwidth}{!}{

\begin{tikzpicture}
  \node[obs]                                              (x1) {$x_1$};
  \node[obs, right=1.6 of x1]                               (x2) {$x_2$};
  \node[obs, right=1.6 of x2]                               (x3) {$x_3$};
  \node[obs, right=1.6 of x3]                               (xt) {$x_t$};
  \node[obs, right=3.2 of xt]                               (xT) {$x_T$};

  \node[latent, above=1 of x1]                           (h1) {$h_1$};
  \node[latent, right=1.6 of h1]                           (h2) {$h_2$};
  \node[latent, right=1.6 of h2]                           (h3) {$h_3$};
  \node[latent, right=1.6 of h3]                           (ht) {$h_t$};
  \node[latent, right=3.2 of ht]                           (hT) {$h_T$};

  \node[latent, above=0.1 of h1]                         (z1) {$\zeta_1$};
  \node[latent, right=1.6 of z1]                           (z2) {$\zeta_2$};
  \node[latent, right=1.6 of z2]                           (z3) {$\zeta_3$};
  \node[latent, right=1.6 of z3]                           (zt) {$\zeta_t$};
  \node[latent, right=3.2 of zt]                           (zT) {$\zeta_T$};

  \node[obs, above=0.1 of z1]                         (y1) {$y$};
  \node[obs, right=1.6 of y1]                           (y2) {$y$};
  \node[obs, right=1.6 of y2]                           (y3) {$y$};
  \node[obs, right=1.6 of y3]                           (yt) {$y$};
  \node[obs, right=3.2 of yt]                           (yT) {$y$};


  \plate {t1} {(h1)(z1)(y1)} {} 
  \plate {t2} {(h2)(z2)(y2)} {} 
  \plate {t3} {(h3)(z3)(y3)} {} 
  \plate {tt} {(ht)(zt)(yt)} {} 
  \plate {tT} {(hT)(zT)(yT)} {} 
  
  \factor[below=0.25 of t1] {} {right:$\Psi_*^{N}(\mathbf{x},h,y)$} {t1,x1} {} ; %
  \factor[below=0.25 of t2] {} {} {t2,x2} {} ; %
  \factor[below=0.25 of t3] {} {} {t3,x3} {} ; %
  \factor[below=0.25 of tt] {} {} {tt,xt} {} ; %
  \factor[below=0.25 of tT] {} {} {tT,xT} {} ; %

  \factor[right=0.6 of t1] {} {} {t1,t2} {} ; %
  \factor[right=0.6 of t2] {} {} {t2,t3} {} ; %
  \factor[right=0.6 of t3] {} {} {t3,tt} {} ; %
  \factor[right=1.3 of tt] {ft} {} {tt,tT} {} ; %
 
  \node[const,above=1.4 of ft] (name) {$\Psi_*^E(h_t,h_{t+1},\zeta_t,\zeta_{t+1},y)$};
 
  \edge[->] {name} {ft} ; %

  \end{tikzpicture}
} \\
        {(a)}  & ~ & ~ {(b)} \\
    \end{tabular}
    \caption{{(a) Factor graph representation of the proposed MI-DORF framework. Node potentials $\Psi^{N}$ model the compatibility between a given observation $\mathbf{x}_t$ and a latent ordinal value $h_t$ . Edge potentials $\Psi^E$ take into account the transition between consecutive latent ordinal states $h_t$ and $h_{t+1}$. Finally, the high-order potential $\Psi^{M}$ models Multi-Instance assumptions relating all the latent ordinal states $\mathbf{h}_t$  with the bag-label $y$. (b) Equivalent model to MI-DORF defined using the auxiliary variables $\zeta_t$ for each latent ordinal state. The use of these auxiliary variables and the redefinition of node and edge potentials allows to perform efficient inference by removing the high-order dependency introduced by the potential $\Psi^{M}$ (see Sec. \ref{sec:inferenceMMI-DORF} and \ref{sec:inferenceRMI-DORF}).}}
    \label{fig:mildorf}
\end{figure*}
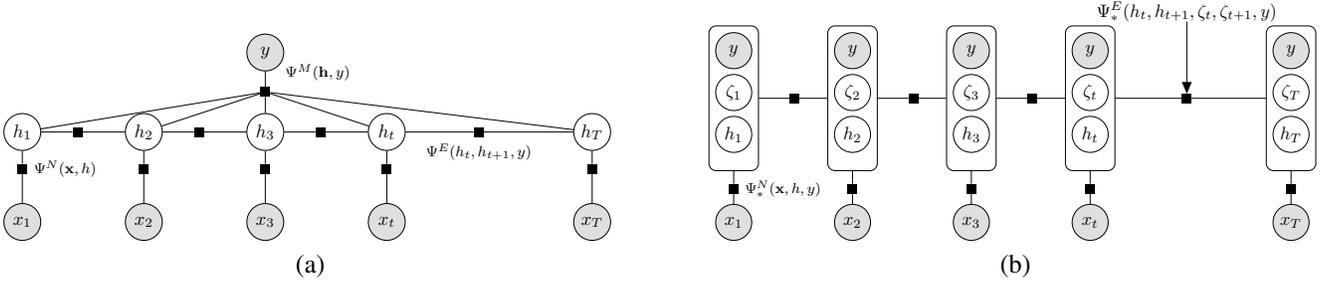

\section{Multi-Instance Dynamic Ordinal Regression}
\label{sec:MIDOR}
In this section, we formalize the MI-DOR problem and its particular instances addressed in this work: Maximum MI-DOR and Relative MI-DOR. In these tasks we are provided with a training set $\mathcal{T}=\{(\mathbf{X}_1,y_1),(\mathbf{X}_2,y_2),...,(\mathbf{X}_N,y_N)\}$ formed by pairs of structured-inputs  $X\in\mathcal{X}$ and labels $y$. Specifically,  $\mathbf{X} = \{ \mathbf{x}_{1},\mathbf{x}_{2},...,\mathbf{x}_{T} \}$ are temporal sequences of $T$ observations $\mathbf{x} \in R^d$ in a d-dimensional space.  Given the training-set $\mathcal{T}$, the goal is to learn a model $\mathcal{F}: \mathcal{X} \rightarrow \mathcal{H}$ mapping sequences $\mathbf{X}$ to an structured-output $\mathbf{h} \in \mathcal{H}$. Concretely, $\mathbf{h} = \{ h_{1},h_{2},...,h_{T} \}$ is a sequence of variables $h_t \in \{ 0 \prec ... \prec l \prec L \}$ assigning one ordinal value for each observation $\mathbf{x_t}$. In order to learn the model $\mathcal{F}$ from $\mathcal{T}$, it is necessary to incorporate prior knowledge defining the Multi-Instance relation between labels $y$ and latent ordinal states $\mathbf{h}$.  In Maximum MI-DOR, we assume that bag-labels $y\in \{ 0 \prec ... \prec l \prec L \}$ are also ordinal variables and that the maximum value in $\mathbf{h}_n$ must be equal to $y_n$:

\begin{equation}
\label{eq:miorassumption}
y_n = \max_h(\mathbf{h}_n) \hspace{4mm} \forall \hspace{1mm} (\mathbf{X}_n,y_n) \in  \mathcal{T}
\end{equation}

On the other hand, in Relative MI-DOR the sequence label is a categorical variable taking four possible values $y \in \{ \uparrow,\downarrow,\oldemptyset,\updownarrow \}$. Each label indicates the type of evolution within latent labels $\mathbf{h}$. Concretely, in sequences labelled with $y=\uparrow$, there must be an increasing ordinal level transition in, at least, one instant $t$. Moreover, no decreasing transitions are allowed within the sequence. The opposite occurs in sequences labelled as $y=\downarrow$. In the case of $y=\updownarrow$ the sequence is assumed to contain decreasing and increasing transitions. Finally, when $y=\oldemptyset$ all the ordinal values in $\mathbf{h}$ should be equal (monotone sequence). Formally, these constraints can be defined as:

\begin{align}
 \forall \hspace{1mm} (\mathbf{X}_n,y_n)
 \begin{cases}  y_n = \uparrow \hspace{3mm} \text{iff} \hspace{3mm}  (\exists t \hspace{1mm} h_t<h_{t+1})  \land (\forall t \hspace{1mm} h_t \leq h_{t+1}) \\
    y_n = \downarrow \hspace{3mm} \text{iff} \hspace{3mm} (\exists t \hspace{1mm} h_t>h_{t+1}) \land (\forall t \hspace{1mm} h_t \geq h_{t+1})  \\
    y_n = \oldemptyset \hspace{1mm} \text{iff} \hspace{3mm} (\forall t \hspace{1mm} h_t = h_{t+1})   \\
    y_n = \updownarrow \hspace{3mm} \text{otherwise}
  \end{cases}
  \label{eq:relmiorassumption}
\end{align}

Note that the definition of these MI-DOR problems differs from standard supervised sequence classification with latent variables. In that case, the main goal is to learn a model $\mathcal{F}: \mathcal{X} \rightarrow \mathcal{Y}$ mapping $\mathbf{X}$ to sequence labels $y$.


\section{Max-Multi-Instance Dynamic Odinal Random Fields (MaxMI-DORF)}
\label{sec:MMIDORF}

In this section, we present the proposed Max-Multi-Instance Dynamic Ordinal Random Fields to solve the Maximum MI-DOR problem described in Sec. \ref{sec:MIDOR}. 

\subsection{Model Definition}

MaxMI-DORF is an Undirected Graphical Model defining the conditional probability of labels $y$ given observations $\mathbf{X}$ with a Gibbs distribution:

\begin{equation}
\label{eq:cond_probability}
P(y|\mathbf{X};\theta) = \sum_\mathbf{h} P(y,\mathbf{h} |\mathbf{X};\theta) =\frac{\sum_h{e^{-\Psi(\mathbf{X},\mathbf{h},y;\theta)}}}{\sum_{y^\prime}\sum_h{e^{-\Psi(\mathbf{X},\mathbf{h},y^\prime;\theta)}}},
\end{equation}
where $\theta$ is the set of the model parameters and the energy function $ \Psi(\mathbf{X},\mathbf{h},y;\theta)$  is composed of the sum of three different types of potentials (see Fig. \ref{fig:mildorf}(a)):

\begin{equation}
\label{eq:energy_func}
 \sum_{t=1}^{T} \Psi^{N}(\mathbf{x}_t,h_t;\theta^N)
  + \sum_{t=1}^{T-1} \Psi^{E}(h_t,h_{t+1};\theta^E)  
  + \Psi^{M}(\mathbf{h},y,\theta^{M}),
\end{equation}

\subsubsection{MaxMI-DORF: Ordinal node potential} This potential $\Psi^{N}(\mathbf{x},h;\theta^N)$ aims to capture the compatibility between a given observation $\mathbf{x}_t$ and the latent ordinal value $h_t$. Similar to HCORF, it is defined using the ordered probit model \cite{winkelmann2006analysis}:

\begin{equation}
  \label{eq:ordinal_regression}
  \Psi^{N}(\mathbf{x},h=l)= \log \Bigg[  \Phi \bigg( \frac{b_l - \mathbf{\beta}^T \mathbf{x}}{\sigma}\bigg) - \Phi \bigg( \frac{b_{(l-1)} - \mathbf{\beta}^T \mathbf{x}}{\sigma}\bigg) \Bigg], 
\end{equation}
where $\Phi(\cdot)$ is the normal cumulative distribution function (CDF), and $\theta^N=\{\beta,\mathbf{b},\sigma\}$ is the set of potential parameters. Specifically, the vector $\beta \in {R}^d$ projects observations $\mathbf{x}$ onto an ordinal line divided by a set of cut-off points ${b_0} =  - \infty  \le  \cdots  \le {b_L} = \infty $. Every pair of contiguous cut-off points divide the projection values into different bins corresponding to the different ordinal states $l=1,...,L$. The difference between the two CDFs provides the probability of the latent state $l$ given the observation $\mathbf{x}$, where $\sigma$ is the standard deviation of a Gaussian noise contaminating the ideal model (see 
\cite{kim2010hidden} for more details). In our case, we fix $\sigma=1$, to avoid model over-parametrization. {This type of potentials has previously been shown to be effective for Ordinal Regression problems  such as AU or Pain Intensity estimation \cite{rudovic2013automatic,rudovic2015context}.}

\subsubsection{MaxMI-DORF: Edge potentials} The edge potential $\Psi^E(h_t,h_{t+1};\theta^E)$ models temporal information regarding compatibilities between consecutive latent ordinal states as:

\begin{equation}
 \Psi^E(h_t=l,h_{t+1}=l^\prime;\theta^E) = f(\mathbf{W}_{l,l^\prime}),
\end{equation}
where $\theta^E={\mathbf{W}^{L \times L}}$ represents a real-valued transition matrix as in standard HCR. On the other hand, $f$ is a non-linear function defined as $f(s)= -\log(1+exp(-s))$. The motivation of using $f$ is to maintain the same range between the values of node and edge potentials. Specifically, $f$ bounds the value of $\Psi^E$ between $[0,-\infty]$ as in the case of the node potentials.

\subsubsection{MaxMI-DORF: Multi-Instance potential}  In order to model the Maximum MI-DOR assumption (see Eq. \ref{eq:miorassumption}), we define a high-order potential $\Psi^{M}(\mathbf{h},y;\theta^M)$ involving label $y$ and all the sequence latent variables $\mathbf{h}$ as:

\begin{equation}
\label{eq:mil_potential}
 \Psi^{M}(\mathbf{h},y;\theta^M) = 
 \begin{cases}
            w \sum_{t=1}^{T} \mathbf{I} (h_t==y) \hspace{3mm} \text{iff} \hspace{2mm} \max (\mathbf{h}) = y \\
            -\infty \hspace{2mm} \text{otherwise}
    \end{cases},
\end{equation}
where $\mathbf{I}$ is the indicator function, and $\theta^M=w$. Note that when the maximum value within $\mathbf{h}$ is not equal to $y$, the energy function is equal to $-\infty$ and, thus, the probability $P(y|\mathbf{X};\theta)$ drops to 0. On the other hand, if the MI assumption is fulfilled, the summation $w \sum_{t=1}^{T} \mathbf{I} (h_t==y)$ increases the energy proportionally to $w$ and the number of  latent states $\mathbf{h} \in h_t$ that are equal to $y$. This is convenient since, in sequences annotated with a particular label, it is more likely to find many latent ordinal states with such ordinal level. 
Eq. \ref{eq:mil_potential} shares some relations with the cardinality potentials \cite{gupta2007efficient} also employed in binary Multi-Instance Classification \cite{hajimirsadeghi2013multiple}.

\subsection{MaxMI-DORF: Learning}
\label{sec:training}

Given a training set $\mathcal{T}$, we learn the model parameters $\theta$ by minimizing the regularized log-likelihood:
\begin{equation}
\label{eq:training}
 \min_\mathbf{\theta} \hspace{3mm} \sum_{i=1}^N \log P(y|\mathbf{X};\theta) + \mathcal{R}(\theta),
\end{equation}
where the regularization function $\mathcal{R}(\theta)$ over the model parameters is defined as:
\begin{equation}
 \mathcal{R}(\theta) = \alpha (||\beta||_2^2 + ||\mathbf{W}||_F^2)
\end{equation}
and $\alpha$ is set via a validation procedure. {We use L2 regularization because, in related Latent Variable models such as HCRF \cite{quattoni2007hidden} or HCORF \cite{kim2010hidden}, it has been shown to provide an effective mechanism to reduce overfitting.}

The objective function in Eq.\ref{eq:training} is differentiable and standard gradient descent methods can be applied for optimization. To this end, we use the L-BFGS Quasi-Newton method \cite{byrd1994representations}. The gradient evaluation involves marginal probabilities $p(h_t|\mathbf{X})$ and $p(h_t,h_{t+1}|\mathbf{X})$ which can be efficiently computed using the proposed algorithm in Sec. \ref{sec:inferenceMMI-DORF}.


\subsection{MaxMI-DORF: Inference}
\label{sec:inferenceMMI-DORF}

The evaluation of the conditional probability $P(y|\mathbf{X};\theta)$ in Eq.\ref{eq:cond_probability} requires computing $\sum_h{e^{-\Psi(\mathbf{X},\mathbf{h},y;\theta)}}$ for each label $y$. Given the exponential number of possible latent states $\mathbf{h} \in \mathcal{H}$, efficient inference algorithms need to be used. In the case of  Latent-Dynamic Models such as HCRF/HCORF, the forward-backward algorithm \cite{barber2012bayesian} can be applied. This is because the pair-wise linear-chain connectivity between latent states $\mathbf{h}$. However, in the case of MaxMI-DORF, the inclusion of the MIL potential $\Psi^{M}(\mathbf{h},y;\theta^{M})$ introduces a high-order dependence between the label $y$ and all the latent states in $\mathbf{h}$. Inference methods with cardinality potentials have been previously proposed in \cite{gupta2007efficient,tarlow2012fast}. However, these algorithms only consider the case where latent variables are independent and, therefore, they can not be applied in our case. For these reasons, we propose an specific inference method. The idea behind it is to apply the standard forward-backward algorithm by converting the energy function defined in Eq. \ref{eq:energy_func} into an equivalent one preserving the linear-chain connectivity between latent states $\mathbf{h}$.  

To this end, we introduce a new set of auxiliary variables $\boldsymbol{\zeta} = \{\zeta_1,\zeta_2,...,\zeta_T\}$, where each $\zeta_t \in \{0,1\}$ takes a binary value denoting whether the sub-sequence $\mathbf{h}_{1:t}$ contains at least one ordinal state $h$ equal to $y$. Now we define an alternative MaxMI-DORF energy function $\Psi_*$ as:
\begin{align}
\label{eq:energy_func_inference}
 \Psi_*(\mathbf{X},\mathbf{h},\boldsymbol{\zeta},y;\theta) &= \sum_{t=1}^{T} \Psi_*^{N}(\mathbf{x}_t,h_t,\zeta_t,y;\theta^N) \\ \nonumber
 & + \sum_{t=1}^{T-1} \Psi_*^{E}(h_t,h_{t+1},\zeta_t,\zeta_{t+1},y;\theta^E),
\end{align}

where the new node pontentials $\Psi_*^{N}$ and edge potentials $\Psi_*^{E}$ are given by:
 \begin{equation}
  \label{eq:inference_node_potential}
  \Psi_*^{N} = 
   \begin{cases}
            \Psi^{N}(\mathbf{x}_t,h_t;\theta^N)+ w\mathbf{I} (h_t=y) \hspace{1mm} \text{iff} \hspace{1mm} h_t <= y \\
            -\infty \hspace{3mm} \text{otherwise} 
    \end{cases} \nonumber
\end{equation}

\begin{align}
 \Psi_*^{E} = \begin{cases}
            \mathbf{W}_{h_t,h_{(t+1)}} \hspace{3mm} \text{iff} \hspace{3mm} \zeta_t=0 \land \zeta_{t+1}=0 \land h_{t+1} \neq y\\
            \mathbf{W}_{h_t,h_{(t+1)}} \hspace{3mm} \text{iff} \hspace{3mm} \zeta_t=0 \land \zeta_{t+1}=1 \land h_{t+1} = y \\
            \mathbf{W}_{h_t,h_{(t+1)}} \hspace{3mm} \text{iff} \hspace{3mm} \zeta_t=1 \land \zeta_{t+1}=1 \\
            -\infty \hspace{3mm} \text{otherwise}
  \end{cases}
  \label{eq:aux_edge_maxmidorf}
\end{align}

 Note that Eq. \ref{eq:energy_func_inference} does not include the potential $\Psi^{M}$, thus, the high-order dependence between the label $y$ and latent ordinal-states $\mathbf{h}$ is removed.  The graphical representation of MI-DORF with the redefined energy function is illustrated in Fig.\ref{fig:mildorf}(b). In order to show the equivalence between energies in Eqs. \ref{eq:energy_func} and \ref{eq:energy_func_inference}, we explain how the the original Multi-Instance potential $\Psi^M$ is incorporated into the new edge and temporal potentials. Firstly, note that $\Psi^{N}$ now also takes into account the proportion of ordinal variables $h_t$ that are equal to the sequence label. Moreover, it enforces $\mathbf{h}$ not to contain any $h_t$ greater than $y$, thus aligning the bag and (max) instance labels. However, the original Multi-Instance potential also constrained $\mathbf{h}$ to contain at least one $h_t$ with the same ordinal value than $y$. This is achieved by using the set of auxiliary variables $\zeta_t$ and the re-defined edge potential $\Psi^{E}$. In this case, transitions between latent ordinal states are modelled but also between auxiliary variables $\zeta_t$. Specifically, when the ordinal state in $h_{t+1}$ is equal to $y$, the sub-sequence $\mathbf{h}_{1:t+1}$ fulfills the Maximum MI-DOR assumption and, thus, $\zeta_{t+1}$ is forced to be $1$. By defining the special cases at the beginning and the end of the sequence ($t=1$ and $t=T$):

 \begin{equation}
  \label{eq:inference_node_potential0}
  \Psi_*^{N}(\mathbf{x}_1,h_1,\zeta_1,y) = 
   \begin{cases}
            \Psi_*^{N} \hspace{2mm} \text{iff} \hspace{2mm} \zeta_1 = 0 \land l_1 < y \\
            \Psi_*^{N} \hspace{2mm} \text{iff} \hspace{2mm} \zeta_1 = 1 \land l_1 = y \\
            -\infty \hspace{3mm} \text{otherwise}
    \end{cases},
\end{equation}

 \begin{equation}
  \label{eq:inference_node_potentialT}
  \Psi_*^{N}(\mathbf{x}_T,h_T,\zeta_T,y) = 
   \begin{cases}
            \Psi_*^{N}  \hspace{1.5mm} \text{iff} \hspace{1.5mm} \zeta_T = 1 \land h_T <= y  \\
            -\infty \hspace{3mm} \text{otherwise}
    \end{cases}
\end{equation}

we can see that the energy is $-\infty$ when the Maximum MI-DOR assumption is not fulfilled. Otherwise, it has the same value than the  one defined in Eq.\ref{eq:energy_func} since no additional information is given. The advantage of using this equivalent energy function is that the standard forward-backward algorithm can be applied to efficiently compute the conditional probability:
\begin{equation}
\label{eq:cond_probability2}
P(y|\mathbf{X};\theta) = \frac{\sum_\mathbf{h} \sum_{\boldsymbol{\zeta}} {e^{-\Psi_*(\mathbf{X},\mathbf{h},\boldsymbol{\zeta},y;\theta)}}}{\sum_{y^\prime}\sum_\mathbf{h} \sum_{\boldsymbol{\zeta}} {e^{-\Psi_*(\mathbf{X},\mathbf{h},\boldsymbol{\zeta},y^\prime;\theta)}}},
\end{equation}

The proposed procedure has a computational complexity of $\mathcal{O}(T \cdot (2L)^2)$ compared with $\mathcal{O}(T \cdot L^2)$ using standard forward-backward in traditional linear-chain latent dynamical models. Since typically $L<<T$, this can be considered a similar theoretical complexity. The presented algorithm can also be applied to compute the marginal probabilities $p(h_t|\mathbf{X})$ and $p(h_t,h_{t+1}|\mathbf{X})$. 


\section{Relative-Multi-Instance DORF (RelMI-DORF)}
\label{sec:RMIDORF}

In this section, we present the proposed Relative-Multi-Instance Dynamic Odinal Random Fields to solve the Relative MI-DOR problem described in Sec. \ref{sec:MIDOR}. 

\subsection{RelMI-DORF: Model Definition}

In RelMI-DORF, ordinal and node potentials are specified as in MaxMi-DORF. However, the Multi-Instance potential $\Psi^{M}(\mathbf{h},y)$ it is now defined as shown in Eq. \ref{eq:mil_potentialRel}. In this case, the potential models the Relaltive MI-DOR assumption, i.e, the weak-relation between the sequence label $y$ and the evolution of latent instance labels $\mathbf{h }$ (see Eq. \ref{eq:relmiorassumption}).

\begin{equation}
\label{eq:mil_potentialRel}
 \Psi^{M} = 
 \begin{cases}
            0 \hspace{2mm} \text{iff} \hspace{2mm} (\exists \hspace{1mm} t h_t<h_{t+1}) \hspace{1mm} \land \hspace{1mm} (\forall t \hspace{1mm} h_t \leq h_{t+1}) \hspace{1mm} \land \hspace{1mm} y = \uparrow \\
            0 \hspace{2mm} \text{iff} \hspace{2mm} (\exists t \hspace{1mm} h_t>h_{t+1}) \hspace{1mm} \land \hspace{1mm} (\forall t \hspace{1mm} h_t \geq h_{t+1}) \hspace{1mm} \land \hspace{1mm} y = \downarrow \\
            0 \hspace{2mm} \text{iff} \hspace{2mm} (\exists t \hspace{1mm} h_t>h_{t+1}) \hspace{1mm} \land \hspace{1mm} (\exists t \hspace{1mm} h_t < h_{t+1}) \hspace{1mm} \land\hspace{1mm}  y = \updownarrow \\
            0 \hspace{2mm} \text{iff} \hspace{2mm} (\forall t \hspace{1mm} h_t = h_{t+1}) \hspace{1mm} \land  \hspace{1mm} y = \oldemptyset \\
            -\infty \hspace{2mm} \text{otherwise}
    \end{cases}
\end{equation}

Learning in RelMI-DORF can be performed following the same procedure described in Sec. \ref{sec:training}. However, inference requires a special treatment which is described  as follows.

\subsection{RelMI-DORF: Inference}
\label{sec:inferenceRMI-DORF}
Similar to the case of MaxMI-DORF, the high-order potential $\Psi^N(\mathbf{h},y)$ in RelMI-DORF prevents to perform inference using the standard forward-backward procedure. For this purpose, we follow a similar strategy than the one described in Sec. \ref{sec:inferenceMMI-DORF}. However, in this case, auxiliary variables $\zeta_t$ are defined according to the possible sequence labels in Relative MI-DOR. Concretely, $\zeta_t \in \{ \uparrow,\downarrow,\oldemptyset,\updownarrow \}$ indicates the label of the subsequence $\mathbf{h}_{1:t}$ according to the definitions given in Eq. \ref{eq:relmiorassumption}. The equivalent energy function incorporating this auxiliary variables $\mathbf{\zeta}$ can be obtained by redefining the original edge potentials as: 

\begin{align}
 \Psi_*^{E} = \begin{cases}
            \mathbf{W}_{h_t,h_{(t+1)}} \hspace{3mm} \text{iff} \hspace{3mm} \zeta_t=\oldemptyset \hspace{1mm} \land \hspace{1mm} \zeta_{t+1}=\oldemptyset \hspace{1mm} \land \hspace{1mm} h_{t} = h_{t+1}\\
            \mathbf{W}_{h_t,h_{(t+1)}} \hspace{3mm} \text{iff} \hspace{3mm} \zeta_t=\oldemptyset \hspace{1mm} \land \hspace{1mm} \zeta_{t+1}= \hspace{0.9mm}\uparrow \hspace{1mm} \land \hspace{1mm} h_{t} < h_{t+1}\\
            \mathbf{W}_{h_t,h_{(t+1)}} \hspace{3mm} \text{iff} \hspace{3mm} \zeta_t=\oldemptyset \hspace{1mm} \land \hspace{1mm} \zeta_{t+1}=\hspace{0.9mm}\downarrow \hspace{1mm} \land \hspace{1mm} h_{t} > h_{t+1}\\
            \mathbf{W}_{h_t,h_{(t+1)}} \hspace{3mm} \text{iff} \hspace{3mm} \zeta_t=\hspace{0.9mm}\uparrow \hspace{1mm} \land \hspace{1mm} \zeta_{t+1}=\hspace{0.9mm}\uparrow \hspace{1mm} \land \hspace{1mm} h_{t} \leq h_{t+1}\\
            \mathbf{W}_{h_t,h_{(t+1)}} \hspace{3mm} \text{iff} \hspace{3mm} \zeta_t=\hspace{0.9mm}\uparrow \hspace{1mm} \land \hspace{1mm} \zeta_{t+1}=\hspace{0.9mm}\updownarrow \hspace{1mm} \land \hspace{1mm} h_{t} > h_{t+1}\\
            \mathbf{W}_{h_t,h_{(t+1)}} \hspace{3mm} \text{iff} \hspace{3mm} \zeta_t=\hspace{0.9mm}\downarrow \hspace{1mm} \land \hspace{1mm} \zeta_{t+1}=\hspace{0.9mm}\downarrow \hspace{1mm} \land \hspace{1mm} h_{t} \geq h_{t+1} \\
            \mathbf{W}_{h_t,h_{(t+1)}} \hspace{3mm} \text{iff} \hspace{3mm} \zeta_t=\hspace{0.9mm}\downarrow \hspace{1mm} \land \hspace{1mm} \zeta_{t+1}=\hspace{0.9mm}\updownarrow \hspace{1mm} \land \hspace{1mm} h_{t} < h_{t+1}\\
            \mathbf{W}_{h_t,h_{(t+1)}} \hspace{3mm} \text{iff} \hspace{3mm} \zeta_t=\hspace{0.9mm}\updownarrow \hspace{1mm} \land \hspace{1mm} \zeta_{t+1}=\hspace{0.9mm}\updownarrow \\     
            -\infty \hspace{3mm} \text{otherwise}
  \end{cases}
  \label{eq:aux_edge_relmidorf}
\end{align}
Again, defining the special cases for node potentials at the beginning and ending of the sequence:
 \begin{equation}
  \label{eq:inference_node_potential0Rel}
  \Psi_*^{N}(\mathbf{x}_1,h_1,\zeta_1,y) = 
   \begin{cases}
            \Psi^{N}(\mathbf{x}_1,h_1,y) \hspace{2mm} \text{iff} \hspace{2mm} \zeta_1 = \oldemptyset \\
            -\infty \hspace{3mm} \text{otherwise}
    \end{cases},
\end{equation}
 \begin{equation}
  \label{eq:inference_node_potentialTRel}
  \Psi_*^{N}(\mathbf{x}_T,h_T,\zeta_T,y) = 
   \begin{cases}
            \Psi^{N}(\mathbf{x}_T,h_T,y)  \hspace{1.5mm} \text{iff} \hspace{1.5mm} \zeta_T = y \\
            -\infty \hspace{3mm} \text{otherwise}
    \end{cases},
\end{equation}

it can be shown that the energy function becomes $-\infty$ when the sequence level is not coherent with the evolution of latent instance labels $\mathbf{h}$ (according to sequence label $y$ and the Relative MI-DOR assumption). Otherwise, it takes the same value than the  energy function defined by the original potentials. In this case, computational complexity  is $\mathcal{O}(T \cdot (4L)^2)$, which is still linear in terms of instances $T$.


\section{Partially-Observed MI-DOR (PoMI-DOR)}
\label{sec:pomidorf}
Although labels at sequence-level are easier to collect, in some applications is feasible to annotate  a small subset of the sequence's instances. In this case, we are interested in learning the model by using weak-labels $y$ but also incorporating the information of these additional annotations. We refer to this problem as Partially-Observed Multi-Instance Dynamic Ordinal Regression (PoMI-DOR). In this case, the training set is formed by triples $\mathcal{T}=\{(\mathbf{X}_1,y_1,\mathbf{h}_1^a),(\mathbf{X}_2,y_2,\mathbf{h}_2^a),...,(\mathbf{X}_N,y_N,\mathbf{h}_N^a)\}$, where $\mathbf{h}_n^a$ contains ground-truth annotations for a subset of sequence instances. Formally, the set $\mathbf{h}_n=\{\mathbf{h}_n^a \cup \mathbf{h}_n^u\}$, where $\mathbf{h}_n^u$ is the subset of ordinal labels corresponding to non annotated instances. 
Under this setting, we extend MI-DORF to learn a model maximizing the log-likelihood function of the conditional probability:

\begin{equation}
\label{eq:cond_probabilityWSSDOR}
P(y,\mathbf{h}_a|\mathbf{X};\theta) = \frac{\sum_{\mathbf{h}^u}{e^{-\Psi(\mathbf{X},\mathbf{h}^u,\mathbf{h}^a,y;\theta)}}}{\sum_{y^\prime}\sum_{\mathbf{h}^u}\sum_{\mathbf{h}^a}{e^{-\Psi(\mathbf{X},\mathbf{h}^u,\mathbf{h}^a,y^\prime;\theta)}}},
\end{equation}
for all the sequences in the training set. Note that in this case, the knowledge provided by annotated instances $\mathbf{h}^a_n$ is incorporated into the likelihood function. In order to learn a PoMI-DORF model , the same algorithms presented in Secs. \ref{sec:MMIDORF} and \ref{sec:RMIDORF} can be applied. However, during inference we need to take into account annotations $\mathbf{h}^a_n$ for each sequence. This can be easily achieved by redefining the original node potentials in RelMI-DORF and MaxMI-DORF as: 

 \begin{equation}
  \label{eq:inference_node_potentialWSSDOR}
  \Psi^{N}(\mathbf{x},h_t) = 
   \begin{cases}
            -\infty \hspace{3mm} \text{iff} \hspace{2mm} (h_t \in \mathbf{h}^a) \land (h^a_t \neq h_t)\\
            \Psi^{N}(\mathbf{x}_t,h_t) \hspace{2mm} \text{otherwise}
    \end{cases},
\end{equation}

Intuitively, observed instance labels $\mathbf{h}^a$ are treated as hard evidences which make the energy function to take a value of $-\infty$ when $\mathbf{h}$ is not consistent with them. This strategy has been previously followed in order to learn Conditional Random Fields \cite{li2009extracting} under the partially-observed setting.


\section{Experiments}


\subsection{Compared methods}
\label{sec:baselines}

The presented frameworks are designed to address Multi-Instance-Learning problems when bags are structured as temporal sequences of instances with ordinal labels. Given that this has not been attempted before, we evaluate alternative methods that can be also used in these problems but present some limitations: either ignore the MIL assumptions (Single-Instance), do not model dynamic information (Static) or do not take into account the ordinal nature of instance labels.

\textbf{Single-Instance Ordinal Regression (SIL-OR):} Maximum MI-DOR can be posed as a supervised learning problem with noisy labels. The main assumption is that the majority of instances will have the same label than their bag. In order to test this assumption, we train standard Ordinal Regression \cite{winkelmann2006analysis} at instance-level by setting all their labels to the same value as their corresponding bag. This baseline can be considered an Static-SIL approach to solve the Maximum MI-DOR problem.

\textbf{Static Multi-Instance Ordinal Regression (MI-OR):} Again for Maximum MI-DOR, we have implemented this Static Multi-Instance approach. This method is inspired by MI-SVM \cite{andrews2002support}, where instance labels are considered latent variables and are iteratively optimized during training. To initialize the parameters of the ordinal regressor, we follow the same procedure as described above in SIL-OR. Then, ordinal values for each instance are predicted and modified so that the Maximum MI-DOR assumption is fulfilled for each bag. 
Ordinal Regression is applied again and this procedure is applied iteratively until convergence. 

\textbf{Multi-Instance-Regression (MIR):} As discussed in Sec. \ref{sec:introduction}, the Maximum MI-DOR problem is closely related with Multiple-Instance-Regression. In order to evaluate the performance of this strategy, we have implemented a similar method as used in \cite{hsu2014augmented}. Note that this approach does not model temporal information and treat ordinal labels as continuous variables.

\textbf{MaxMI-DRF:} This approach is similar to the proposed MaxMI-DORF. However, MaxMI-DRF ignores the ordinal nature of labels and models them as categorical variables. For this purpose, we replace the MaxMI-DORF node potentials by a multinomial logistic regression model \cite{walecki2015variablestate}. 
Inference is performed by using the same algorithm described in Sec. \ref{sec:inferenceMMI-DORF}.

\textbf{RelMI-DRF:} Similar to MaxMI-DRF, this method is equivalent to RelMI-DORF but modelling instance labels as categorical variables.

\textbf{Latent-Dynamic Models (HCRF/HCORF):} In Maximum and Relative MI-DOR a label at sequence-level is provided during training. Therefore, it is possible to apply existing Latent-Dynamic Models such as HCRF \cite{quattoni2007hidden} or HCORF \cite{kim2010hidden} for both problems. Despite these two methods model dynamics and incorporate the information provided by sequence-labels, they do not take into account the Multi-Instance assumptions.

\textbf{Ordinal Support Vector Regression (OSVR):} This method presented in \cite{zhao2016facial} can be applied for Relative MI-DOR. However, it is an Static approach that do not consider dynamic information. Moreover, it models instance labels as continuous variables instead of ordinal. 

\textbf{Methods for Partially-Observable MI-DOR:} In our experiments, we evaluate Max-MIDORF and Rel-MIDORF when some instance labels are also available during  training (see Sec. \ref{sec:pomidorf}). In order to compare their performance under this setting, we evaluate the partially-observed extensions of CRF \cite{li2009extracting} and HCRF \cite{chang2009learning}. Ordinal versions of these two approaches has been also implemented for this work.

\textbf{Methods for Supervised Dynamic Ordinal Regression:} To fully evaluate the performance of methods trained using only weak-labels, we compare the previous described methods with two related fully-supervised models for sequence classification CRF \cite{lafferty2001conditional} and CORF \cite{kim2010structured}. These approaches are learned with complete information (i.e, labels for all the instances ).

\begin{figure}[t]
    \centering
     \includegraphics[width=0.4\textwidth]{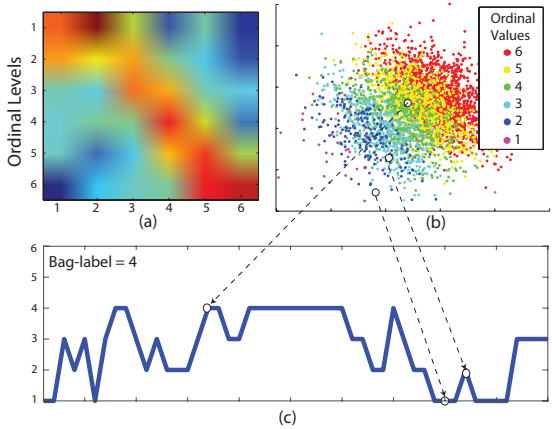}
     \caption{ {Description of the procedure used to generate synthetic sequences. (a) A random matrix modelling transition probabilities between consecutive latent ordinal values. (b) Ordinal levels assigned to the random feature vectors according to the ordinal regressor. (c) Example of a sequence of ordinal values obtained using the generated transition matrix. The feature vector representing each observation is randomly chosen between the samples in (b) according to the probability for each ordinal level.}}
     \label{fig:unbc_results_synthetic}
\end{figure}


\subsection{Metrics and Evaluation}

In order to evaluate the performance of the different methods, we report results in terms of instance-labels predictions. Note that in the MIL literature, results are usually reported at bag-level. However, in MI-DOR problems, the only goal is to predict instance labels (pain or AU intensities) inside the bag (video). Given the ordinal nature of the labels, we use  Pearson's Correlation (CORR), Mean-Average-Error (MAE) and Intra-Class-Correlation (ICC) as evaluation metrics. {In all our experiments, we used a subset of the training sequences to optimize the different regularization weights (hyper-parameters) in a cross-validation procedure. To this end, we used standard grid-search where regularization parameters has been chosen between different values in the range  $[10^{-4},\dots,10^{-1}]$.}




\subsection{Maximum MI-DOR and Relative MI-DOR: Synthetic Data}

\subsubsection{Synthetic Data generation} Given that no standard benchmarks are available for MI-DOR problems, we have generated synthetic data. In order to create sequences for Maximum MI-DOR, we firstly sample a sequence of ordinal values using a random transition matrix representing transition probabilities between temporally-consecutive ordinal levels.  Secondly, we generate random parameters of an Ordinal Regressor as defined in Eq. \ref{eq:ordinal_regression}. This regressor is used to compute the probabilities for each ordinal level in a set of feature-vectors randomly sampled from a Gaussian distribution. Thirdly, the corresponding sequence observation for each latent state in the sequence is randomly chosen between the sampled feature vectors according to the obtained probability for each ordinal value. Finally, the sequence-label is set to the maximum ordinal state within the sequence following the Maximum MI-DOR assumption and Gaussian noise ($\sigma=0.25$) is added to the feature vectors. Fig. \ref{fig:unbc_results_synthetic}(a-c) illustrates this procedure. 

For Relative MI-DOR, we follow a similar strategy to generate the synthetic sequences. However, the transition matrix is forced to contain a probability of 0 for decreasing transitions in case the sequence label is $y=\uparrow$ and for increasing transitions if $y=\downarrow$. For testing, we create unsegmented sequences (with increasing and decreasing transitions) by concatenating two segments generated following the previous procedure. 

\subsubsection{Experimental setup and results} Following the strategy described above, we have generated  ten different data sets for Relative and Maximum MI-DOR by varying the ordinal regressor parameters and transition matrix. Specifically, each dataset is composed of 100 sequences for training, 150 for testing and 50 for validation. The sequences have a variable length between 50 and 75 instances in Maximum MI-DOR and between 15 and 25 in Relative MI-DOR. The dimensionality of the feature vectors was set to 10 and the number of ordinal values to 6. For partially-observed MI-DOR, we have randomly choose one instance per sequence of which its label is also used during training. Table \ref{tab:synthmax_results} and \ref{tab:synthrel_results} shows the results computed as the average performance over the ten datasets for Maximum and Relative MI-DOR respectively. We also report results for fully-supervised CRF and CORF trained considering all the instance labels.

\subsubsection{Maximum MI-DOR discussion} In the Maximum MI-DOR problem, SIL methods (SIL-OR, HCRF and HCORF) obtain lower performance than their corresponding MIL versions (MI-OR, MaxMI-DRF and MaxMI-DORF) in all the evaluated metrics. This is expected since SIL approaches ignore the Multi-Instance assumption. Moreover, HCORF and MaxMI-DORF obtain better performance compared to HCRF and MaxMI-DRF. This is because the former model instance labels as nominal variables, thus, ignoring their ordinal nature. Finally, note that MaxMI-DORF outperforms the static methods MI-OR and MIR. Although these approaches use the Multi-Instance assumption and incorporate the labels ordering, they do not take into account temporal information. In contrast, MaxMI-DORF is able to model the dynamics of latent ordinal states and use this information to make better predictions when sequence observations are noisy.

Looking into the results achieved by the different methods in the PoMI-DOR setting, we can derive the following conclusions. Firstly, HCORF and HCRF improve their performance by taking into account the additional information provided by instance labels. However, we can observe that, under this setting, CRF and CORF obtain lower results than HCORF and HCRF. This is because the later are able to use the sequence-label information together with the provided by labelled instances. Secondly, observe that MaxMI-DRF and MaxMI-DORF still achieves better performance than methods that do not consider the MIL assumption (CORF, CRF, HCRF and HCORF). This shows the importance of explicitly incorporate the Maximum MI-DOR assumption in the model even though instance labels can be available during training. Finally, note that MaxMI-DORF obtain again the best performance, even close to fully-supervised CRF and CORF. This suggest that the need of annotated instances is highly-reduced if the sequence weak-labels are used during learning.


\begin{table}[t]
\centering
\caption{Results on Synthtic Data (MaxMI-DOR)}
\resizebox{.43\textwidth}{!}{
\begin{threeparttable}

\begin{tabular}{
>{ }c 
>{ }c 
>{ }c 
>{ }c 
>{ }c }
\multicolumn{1}{c}{ \textbf{Setting}} & \multicolumn{1}{c}{ \textbf{Method}} & \multicolumn{1}{c}{ \textbf{CORR $\uparrow$}} & \multicolumn{1}{c}{ \textbf{MAE $\downarrow$}} & \multicolumn{1}{l}{ \textbf{ICC $\uparrow$}} \\ \hline
\multicolumn{1}{c|}{ } & \multicolumn{1}{c|}{ \textbf{SI-OR}} & 0.79 & 1.31 & 0.46 \\
\multicolumn{1}{c|}{ } & \multicolumn{1}{c|}{ \textbf{MI-OR}} & 0.82 & 0.62 & 0.70 \\
\multicolumn{1}{c|}{ } & \multicolumn{1}{c|}{ \textbf{HCRF \cite{quattoni2007hidden}}} & 0.05 & 1.99 & 0.05 \\
\multicolumn{1}{c|}{ } & \multicolumn{1}{c|}{ \textbf{HCORF \cite{kim2010hidden}}} & 0.73 & 0.74 & 0.65 \\
\multicolumn{1}{c|}{ } & \multicolumn{1}{c|}{ \textbf{MIR \cite{hsu2014augmented}}} & 0.79 & 0.65 & 0.69 \\
\multicolumn{1}{c|}{ } & \multicolumn{1}{c|}{ \textbf{MaxMI-DRF}} & 0.77 & 0.77 & 0.71 \\
\multicolumn{1}{c|}{\multirow{-7}{*}{ \textbf{MaxMI-DOR}}} & \multicolumn{1}{c|}{ \textbf{MaxMI-DORF}} & \textbf{0.86} & \textbf{0.41} & \textbf{0.85} \\ \hline
\multicolumn{1}{c|}{ } & \multicolumn{1}{c|}{ \textbf{PoCRF \cite{li2009extracting}}} & 0.74 & 0.63 & 0.74 \\
\multicolumn{1}{c|}{ } & \multicolumn{1}{c|}{ \textbf{PoCORF \cite{li2009extracting}*}} & 0.84 & 0.46 & 0.83 \\
\multicolumn{1}{c|}{ } & \multicolumn{1}{c|}{ \textbf{PoHCRF \cite{chang2009learning}}} & 0.79 & 0.57 & 0.78 \\
\multicolumn{1}{c|}{ } & \multicolumn{1}{c|}{ \textbf{PoHCORF \cite{chang2009learning}*}} & 0.86 & 0.42 & 0.85 \\
\multicolumn{1}{c|}{ } & \multicolumn{1}{c|}{ \textbf{MaxMI-DRF}} & 0.82 & 0.52 & 0.81 \\
\multicolumn{1}{c|}{\multirow{-6}{*}{ \textbf{\begin{tabular}[c]{@{}c@{}}PoMaxMI-DOR\\ (1 sample/seq. )\end{tabular}}}} & \multicolumn{1}{c|}{ \textbf{MaxMI-DORF}} & \textbf{0.87} & \textbf{0.38} & \textbf{0.87} \\ \hline
\multicolumn{1}{c|}{ } & \multicolumn{1}{c|}{ \textbf{CRF \cite{lafferty2001conditional}}} & 0.88 & 0.35 & 0.88 \\
\multicolumn{1}{c|}{\multirow{-2}{*}{ \textbf{Supervised DOR}}} & \multicolumn{1}{c|}{ \textbf{CORF \cite{kim2010structured}}} & \textbf{0.89} & \textbf{0.35} & \textbf{0.88}
\end{tabular}
\begin{tablenotes}
(*)Indicates a nominal method that we have extended to the ordinal case.
\end{tablenotes}
\end{threeparttable}
}
\label{tab:synthmax_results}
\end{table}


\begin{table}[t]
\centering
\caption{Results on Synthtic Data (RelMI-DOR)}
\resizebox{.43\textwidth}{!}{
\begin{threeparttable}
\begin{tabular}{cc|ccc}
\textbf{Setting} & \textbf{Method} & \textbf{CORR $\uparrow$} & \textbf{MAE $\downarrow$} & \textbf{ICC $\uparrow$} \\ \hline
\multicolumn{1}{c|}{\multirow{5}{*}{\textbf{RelMI-DOR}}} & \textbf{HCRF \cite{quattoni2007hidden}} & 0.36 & 1.82 & 0.32 \\
\multicolumn{1}{c|}{} & \textbf{HCORF \cite{kim2010hidden}} & 0.85 & 1.32 & 0.80 \\
\multicolumn{1}{c|}{} & \textbf{OSVR \cite{zhao2016facial}} & 0.87 & 3.51 & 0.10 \\
\multicolumn{1}{c|}{} & \textbf{RelMI-DRF} & 0.77 & 1.36 & 0.49 \\
\multicolumn{1}{c|}{} & \textbf{RelMI-DORF} & \textbf{0.89} & \textbf{0.74} & \textbf{0.84} \\ \hline
\multicolumn{1}{c|}{\multirow{7}{*}{\textbf{\begin{tabular}[c]{@{}c@{}}PoRelMI-DOR\\ ( 1 sample/seq. )\end{tabular}}}} & \textbf{PoCRF \cite{li2009extracting}} & 0.82 & 0.64 & 0.81 \\
\multicolumn{1}{c|}{} & \textbf{PoCORF \cite{li2009extracting}*} & 0.89 & 0.43 & 0.89 \\
\multicolumn{1}{c|}{} & \textbf{PoHCRF \cite{chang2009learning}} & 0.83 & 0.60 & 0.83 \\
\multicolumn{1}{c|}{} & \textbf{PoHCORF \cite{chang2009learning}*} & 0.89 & 0.44 & 0.88 \\
\multicolumn{1}{c|}{} & \textbf{OSVR \cite{zhao2016facial}} & 0.87 & 0.61 & 0.85 \\
\multicolumn{1}{c|}{} & \textbf{RelMI-DRF} & 0.88 & 0.49 & 0.87 \\
\multicolumn{1}{c|}{} & \textbf{RelMI-DORF} & \textbf{0.92} & \textbf{0.36} & \textbf{0.91} \\ \hline
\multicolumn{1}{c|}{\multirow{2}{*}{\textbf{Supervised DOR}}} & \textbf{CRF \cite{lafferty2001conditional}} & 0.93 & 0.31 & 0.93 \\
\multicolumn{1}{c|}{} & \textbf{CORF \cite{kim2010structured}} & \textbf{0.93} & \textbf{0.29} & \textbf{0.93}
\end{tabular}
\begin{tablenotes}
(*)Indicates a nominal method that we have extended to the ordinal case.
\end{tablenotes}
\end{threeparttable}
}
\label{tab:synthrel_results}
\end{table}

\subsubsection{Relative MI-DOR discussion} In the Relative MI-DOR problem, we observe similar results as in  Maximum MI-DOR. Firstly, note that non-ordinal approaches (HCRF and RelMI-DRF) obtain the worst performance in most cases. Secondly, RelMI-DORF obtain better performance than HCORF by explicitly modelling the Multi-Instance-Assumption. Finally, OSVR achieves a competitive performance in terms of correlation compared with RelMI-DORF. However, it obtains poor results in terms of MAE and ICC. As discussed in Sec. \ref{sec:related}, OSVR considers labels as continuous variables and do not explicitly model the Relative MI-DOR assumption. Instead, it only ranks the instance labels within the sequence. Therefore, it fails to  estimate the actual scale of the predicted values. 

When  some instance labels are provided (PoRel-MIDOR), all the methods improve their performance by exploiting this additional information. However, the improvement in terms of MAE and ICC is much higher than for correlation. This is because in Relative MI-DOR, sequence labels only provide information  about the evolution of instance labels within the sequence. Therefore, models can achieve a good performance predicting sequence-labels even though the ordinal levels are not accurate. In contrast, when some instance labels are incorporated during training, a better estimation of the ordinal levels can be achieved.  
Finally, note that RelMI-DORF under the PoRelMI-DOR setting achieves again competitive performance compared to fully-supervised CRF and CORF.

\begin{figure*}[ht]
     \centering
     \includegraphics[width=0.95\textwidth]{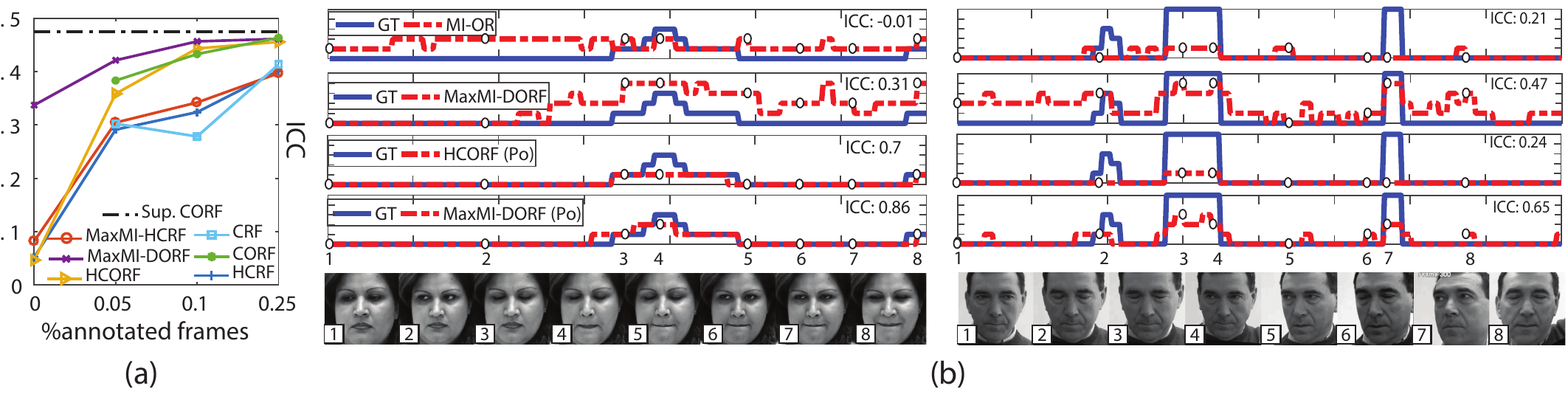}
     \caption{{(a) ICC obtained on the UNBC data when using different percentages of labelled instances from the training set. Black line shows the performance of a fully-supervised CORF trained with all the instance labels. (b) Visualization of the pain intensity predictions in different sequences of the UNBC dataset. From top to bottom: MI-OR and MaxMI-DORF without using instance labels. Partially-observed HCORF and MaxMI-DORF using 10$\%$ of annotated frames.} }
     \label{fig:unbc_qual}
\end{figure*}
\label{sec:syntheicMax}

{\subsubsection{Computational Cost} In order to show the efficiency of the proposed inference algorithms for MaxMI-DORF (Sec. \ref{sec:inferenceMMI-DORF}) and RelMI-DORF (Sec. \ref{sec:inferenceRMI-DORF}), we have computed the average time required to process the testing sequences in each of the 10 synthetic datasets used in our experiments~\footnote{Average computed over 50 different runs for each dataset. Experiment performed using a MATLAB implementation over a Desktop PC (Intel Core i7-4790K@4.00Ghz processor).}. Comparing it with the time required by the  forward-backward procedure employed in HCRF and HCORF, MaxMI-DORF is only 1.6 times slower (0.12s vs. 0.08s). Similarly, the forward-backward algorithm is only 1.5 times faster than RelMI-DORF (0.10s vs. 0.07s). Note that the efficiency of the proposed algorithms is better than expected according to our theoretical analysis. This is because our implementation has been optimized  by exploiting the inherent sparsity of auxiliary node and edge potentials ($-\infty$ cases in Eq.~\ref{eq:aux_edge_maxmidorf} and Eq.~\ref{eq:aux_edge_relmidorf}).}

\subsection{Weakly-supervised pain intensity estimation}
\label{sec:pain_exp}

In this experiment, we test the performance of MaxMI-DORF for weakly-supervised pain intensity estimation. As detailed in Sec. \ref{sec:motivation}, our main motivation is that pain intensity labelling is very time consuming. However, the maximum pain felt during a sequence is much easier to annotate. 

\subsubsection{UNBC Dataset} We use the UNBC Shoulder-Pain Database \cite{lucey2011painful} which contains recordings of different subjects performing active and passive arm movements during rehabilitation sessions. In this dataset, pain intensities at each frame are given in terms of the PSPI scale \cite{prkachin1992consistency}. This ordinal scale ranges from 0 to 15. Given the imbalance between low and high pain intensity levels, we follow the same strategy than \cite{rudovic2015context}. Specifically, pain labels are grouped into 5 ordinal levels as: 0(0),1(1),2(2),3(3),4-5(4),6-15(5). These frame-by-frame pain annotations are considered the instance labels in Maximum MI-DOR. On the other hand, bag (video) labels are extracted as the maximum pain level within each sequence. 

In order to extract facial-descriptors at each video frame representing the bag instances, we compute a geometry-based facial-descriptor as follows. Firstly, we obtain a set of 49 landmark facial-points with the method described in \cite{XiongD13}. Then, the obtained points are aligned with a mean-shape using Procrustes Analysis. Finally, the facial descriptor are obtained by concatenating the $x$ and $y$ coordinates of the aligned points. 
\label{sec:unbc_dataset}

\subsubsection{Experimental setup and results} 
Similar to the experiment with synthetic data (Sec. \ref{sec:syntheicMax}), we consider two scenarios for weakly-supervised pain intensity estimation. The first one is the Maximum MI-DOR setting, where only bag labels are used. Apart from the baselines described in Sec. \ref{sec:baselines}, in this scenario we also evaluate the performance of the approach presented in \cite{sikka2013weakly} which considers pain levels as binary variables. For this purpose, we use the MILBoosting \cite{zhang2005multiple} method employed in the cited work and considered videos with a pain label greater than 0 as positive. Given that MI-Classification methods are only able to make binary predictions, we use the output probability as indicator of intensity levels , i.e., the output probability is normalized between 0 and 5.

We also consider the Partially-Observed setting, where  different percentages of annotated frames inside each sequence are also available during training. This  simulates that the time required to annotate the dataset has been significantly reduced by only labelling a small subset of the frames. Concretely, we consider the ${5\%}$ and ${10\%}$  of annotated frames in each sequence. Under these different experimental setups, we perform Leave-One-Subject-Out Cross Validation where, in each cycle, we use 15 subjects for training, 1 for testing and 9 for validation. In order to reduce computational complexity and redundant information between temporal consecutive frames, we have down-sampled the sequences using a time-step of 0.25 seconds. Table \ref{tab:unbc_results} shows the results obtained by the evaluated methods following the described procedure. Results for fully-supervised CRF and CORF are also reported.

\begin{table}[t]
\caption{Results on the UNBC Database}
\centering
\resizebox{0.43\textwidth}{!}{
\begin{threeparttable}

\begin{tabular}{cc|ccc}
  
\multicolumn{1}{c|}{  \textbf{Setting}} & \multicolumn{1}{c|}{  \textbf{Method} } & \multicolumn{1}{c}{ \textbf{CORR $\uparrow$}} & \multicolumn{1}{c}{ \textbf{MAE $\downarrow$}} & \multicolumn{1}{c}{ \textbf{ICC $\uparrow$}} \\ \hline
  
\multicolumn{1}{c|}{ } & \textbf{SI-OR} & 0.22 & 2.20 & 0.08 \\
  
\multicolumn{1}{c|}{ } & \textbf{MI-OR} & 0.29 & 0.84 & 0.27 \\
  
  \multicolumn{1}{c|}{ } & \textbf{MILBoost \cite{zhang2005multiple}} & 0.23 & 2.38 & 0.09 \\

\multicolumn{1}{c|}{ } & \textbf{HCRF \cite{quattoni2007hidden}} & 0.09 & 1.73 & 0.05 \\
  
\multicolumn{1}{c|}{ } & \textbf{HCORF\cite{kim2010hidden}} & 0.06 & 1.23 & 0.05 \\
  
\multicolumn{1}{c|}{ } & \textbf{MIR \cite{hsu2014augmented}} & 0.32 & 1.03 & 0.25 \\
  
\multicolumn{1}{c|}{ } & \textbf{MaxMI-DRF} & 0.16 & 1.96 & 0.08 \\
  
\multicolumn{1}{c|}{\multirow{-7}{*}{ \textbf{MaxMI-DOR}}} & \textbf{MaxMI-DORF} & \textbf{0.36} & \textbf{0.71} & \textbf{0.34} \\ \hline
  
\multicolumn{1}{c|}{ } & \textbf{PoCRF \cite{li2009extracting}} & 0.31 & 0.66 & 0.30 \\
  
\multicolumn{1}{c|}{ } & \textbf{PoCORF \cite{li2009extracting}*} & 0.39 & 0.58 & 0.38 \\
  
\multicolumn{1}{c|}{ } & \textbf{PoHCRF \cite{chang2009learning}} & 0.32 & 0.76 & 0.29 \\
  
\multicolumn{1}{c|}{ } & \textbf{PoHCORF \cite{chang2009learning}*} & 0.38 & 0.68 & 0.36 \\
  
\multicolumn{1}{c|}{ } & \textbf{MaxMI-DRF} & 0.32 & 0.72 & 0.30 \\
  
\multicolumn{1}{c|}{\multirow{-6}{*}{ \textbf{\begin{tabular}[c]{@{}c@{}}PoMaxMI-DOR\\ (5\% of data )\end{tabular}}}} & \textbf{MaxMI-DORF} & \textbf{0.43} & \textbf{0.52} & \textbf{0.42} \\ \hline
\multicolumn{1}{c|}{} & \textbf{PoCRF \cite{li2009extracting}} & 0.29 & 0.65 & 0.28 \\
\multicolumn{1}{c|}{} & \textbf{PoCORF \cite{li2009extracting}*} & 0.44 & 0.55 & 0.43 \\
\multicolumn{1}{c|}{} & \textbf{PoHCRF \cite{chang2009learning}} & 0.34 & 0.63 & 0.32 \\
\multicolumn{1}{c|}{} & \textbf{PoHCORF \cite{chang2009learning}*} & 0.45 & 0.58 & 0.44 \\
\multicolumn{1}{c|}{} & \textbf{MaxMI-DRF} & 0.34 & 0.55 & 0.34 \\
\multicolumn{1}{c|}{\multirow{-6}{*}{\textbf{\begin{tabular}[c]{@{}c@{}}PoMaxMI-DOR\\ (10\% of data )\end{tabular}}}} & \textbf{MaxMI-DORF} & \textbf{0.46} & \textbf{0.51} & \textbf{0.46} \\ \hline
  
\multicolumn{1}{c|}{ } & \textbf{CRF \cite{lafferty2001conditional}} & 0.45 & \textbf{0.50} & 0.44 \\
  
\multicolumn{1}{c|}{\multirow{-2}{*}{ \textbf{Supervised DOR}}} & \textbf{CORF \cite{kim2010structured}} & \textbf{0.48} & 0.56 & \textbf{0.48}
\end{tabular}
\begin{tablenotes}
(*)Indicates a nominal method that we have extended to the ordinal case.
\end{tablenotes}
\end{threeparttable}
}
\label{tab:unbc_results}
\end{table}

\subsubsection{Discussion} 
 By looking into the results in the Maximum MI-DOR setting, we can derive the following conclusions. Firstly, SI approaches ( SI-OR, HCORF and HCRF) obtain worse performance than MI-OR and MIR. Specially, HCORF and HCRF obtain poor results. This is because pain events are typically very sparse in these sequences and most frames have intensity level 0 (neutral). Therefore, the use of the MIL assumption has a critical importance in this problem in order to correctly locate pain frames. Secondly, MIR and MI-OR obtain better results than MaxMI-DRF. This can be explained because the latter consider pain levels as nominal variables and is ignorant of the ordering information of the different pain intensities. Finally, MILBoost trained with binary labels also obtains low performance compared to the MI-OR and MIR. This suggest that current approaches posing weakly-supervised pain detection as a MI-Classification problem are unable to predict accurately the target pain intensities. By contrast, MaxMI-DORF obtains the best performance across all the evaluated metrics. We attribute this to the fact it models the MIL assumption with ordinal variables. Moreover, the improvement of MaxMI-DORF compared to static approaches, such as MI-OR and MIR, suggests that modelling dynamic information is beneficial in this task. 
 
 In the Partially-observed setting, all the methods improve their performance by considering the additional information provided by labelled instances. However, note that approaches modelling the ordinal structure of labels (CORF, HCORF and MaxMI-DORF) still outperforms nominal methods (CRF, HCRF and MaxMI-DRF) under this setting. Moreover, MaxMI-DORF also achieves the best performance with $5\%$ and $10\%$ of labeled frames. Despite the other approaches also consider instance labels, MaxMI-DORF better exploits sequence labels information by explicitly modelling the MIL assumption. It is worth mentioning that considering only 10$\%$ of annotated frames, MaxMI-DORF obtain competitive performance against fully-supervised approaches. Concretely, it outperforms CRF in terms of ICC/CORR and CORF in terms of MAE. This suggest that the effort needed to annotate pain intensity databases, could be highly-reduced using the proposed weakly-supervised framework. In order to give more insights about this issue, Fig. \ref{fig:unbc_qual}(b) shows the performance in terms of ICC as the percentage of annotated frames increases. As we can observe, MaxMI-DORF outperforms other methods with  $0\%$, $5\%$ and $10\%$ of annotated frames. When this percentage increases to $25\%$, the performance of partially-observed CORF, HCORF and MaxMI-DORF is comparable to the achieved by fully-supervised CORF. However, note that labelling $25\%$ of samples does not suppose a significant reduction of the annotation time in a real scenario. 
 
 Finally, in Fig. \ref{fig:unbc_qual}(b) we show qualitative examples comparing predictions of the best evaluated methods under the different settings. When only bag-labels are used for training, MI-OR predictions are less accurate than the obtained by MaxMI-DORF. Moreover, MaxMI-DORF estimates better the actual pain levels in the partially-observed setting, where a small subset of instance labels are used. These predictions are more accurate than the obtained with partially-observed HCORF which does not take into account the MIL assumption. This is reflected by the ICC depicted in the sequences, showing that the proposed MaxMI-DORF method outperforms the competing approaches on target data.

\begin{table}[t]
\centering
\caption{Average performance across AUs on the DISFA dataset.}

\resizebox{0.43\textwidth}{!}{
\begin{threeparttable}
\begin{tabular}{c|c|ccc}
\multicolumn{1}{c}{\textbf{Setting}} & \textbf{Method} & \textbf{CORR $\uparrow$} & \textbf{MAE $\downarrow$} & \textbf{ICC $\uparrow$} \\ \hline
\multicolumn{1}{c|}{\multirow{5}{*}{\textbf{RelMI-DOR}}} & \textbf{HCRF \cite{quattoni2007hidden}} & 0.21 & 2.04 & 0.10 \\
\multicolumn{1}{c|}{} & \textbf{HCORF \cite{kim2010hidden}} & 0.26 & 3.49 & 0.03 \\
\multicolumn{1}{c|}{} & \textbf{OSVR  \cite{zhao2016facial}} & 0.35 & 1.38 & 0.15 \\
\multicolumn{1}{c|}{} & \textbf{RelMI-DRF} & 0.19 & 1.70 & 0.11 \\
\multicolumn{1}{c|}{} & \textbf{RelMI-DORF} & \textbf{0.40} & \textbf{1.13} & \textbf{0.26} \\ \hline
\multicolumn{1}{c|}{\multirow{7}{*}{\textbf{\begin{tabular}[c]{@{}c@{}}PoRelMI-DOR \\ (5\% frames )\end{tabular}}}} & \textbf{PoCRF \cite{li2009extracting}} & 0.33 & 0.55 & 0.29 \\
\multicolumn{1}{c|}{} & \textbf{PoCORF \cite{li2009extracting}*} & 0.37 & 0.57 & 0.32 \\
\multicolumn{1}{c|}{} & \textbf{PoHCRF \cite{chang2009learning}} & 0.34 & 0.59 & 0.30 \\
\multicolumn{1}{c|}{} & \textbf{PoHCORF \cite{chang2009learning}*} & 0.38 & 0.62 & 0.33 \\
\multicolumn{1}{c|}{} & \textbf{OSVR  \cite{zhao2016facial}} & 0.36 & 0.81 & 0.29 \\
\multicolumn{1}{c|}{} & \textbf{RelMI-DRF} & 0.23 & 0.64 & 0.19 \\
\multicolumn{1}{c|}{} & \textbf{RelMI-DORF} & \textbf{0.40} & \textbf{0.51} & \textbf{0.36} \\ \hline
\multicolumn{1}{c|}{\multirow{7}{*}{\textbf{\begin{tabular}[c]{@{}c@{}}PoRelMI-DOR\\  (10\% frames )\end{tabular}}}} & \textbf{PoCRF \cite{li2009extracting}} & 0.36 & 0.50 & 0.32 \\
\multicolumn{1}{c|}{} & \textbf{PoCORF \cite{li2009extracting}*} & 0.39 & 0.56 & 0.33 \\
\multicolumn{1}{c|}{} & \textbf{PoHCRF \cite{chang2009learning}} & 0.38 & 0.57 & 0.34 \\
\multicolumn{1}{c|}{} & \textbf{PoHCORF \cite{chang2009learning}*} & 0.40 & 0.59 & 0.35 \\
\multicolumn{1}{c|}{} & \textbf{OSVR  \cite{zhao2016facial}} & 0.37 & 0.80 & 0.29 \\
\multicolumn{1}{c|}{} & \textbf{RelMI-DRF} & 0.36 & 0.50 & 0.32 \\
\multicolumn{1}{c|}{} & \textbf{RelMI-DORF} & \textbf{0.42} & \textbf{0.48} & \textbf{0.38} \\ \hline
\multicolumn{1}{c|}{\multirow{2}{*}{\textbf{Supervised DOR}}} & \textbf{CRF \cite{lafferty2001conditional}} & 0.39 & \textbf{0.44} & 0.35 \\
\multicolumn{1}{c|}{} & \textbf{CORF \cite{kim2010structured}} & \textbf{0.41} & 0.50 & \textbf{0.37}
\end{tabular}
\begin{tablenotes}
(*)Indicates a nominal method that we have extended to the ordinal case.
\end{tablenotes}
\end{threeparttable}
}
\label{tab:disfa_table}

\end{table}

\subsection{Weakly-supervised AU intensity estimation}
\label{sec:au_exp}
In this section, we test the performance of RelMI-DORF for weakly-supervised Action Unit intensity estimation. Similarly to pain intensity, AU labelling requires a huge effort for expert coders. However, segmenting videos according to the increasing or decreasing evolution of AU intensities (i.e. onset and offset sequences) is less time-consuming.

\subsubsection{DISFA Dataset} 
We employ the DISFA Database \cite{mavadati2013disfa}, which is a popular benchmark for AU intensity estimation.  It contains naturalistic data consisting on 27 annotated sequences of different subjects watching videos eliciting different types of emotions. Specifically frame-by-frame AU intensities are provided for 12 AUs (1,2,4,5,6,9,12,15,17,20,25,26) in a six-point ordinal scale (\textit{neutral}$<$A$<$B$<$C$<$D$<$E). As far as we know, this is the largest available dataset  in terms of the number of  Action Units annotated. Although the UNBC dataset also provides AU intensity annotations for 11 AUs, we found that the number of onset and appex events for each of them is very limited. Therefore, we discard it for this experiments. To the best of our knowledge, no previous works have evaluated DISFA under the weakly-supervised setting. 

The described AU intensities represent the instance labels in our Relative MI-DOR problem. As previously discussed, bags are considered onset and apex sequences where the intensity of a given AU is monotone increasing  ($y=\uparrow$) or decreasing ($y=\downarrow$). These segments has been automatically extracted with an exhaustive search over the whole video using the ground-truth intensity labels at frame-level. 
This procedure simulates that a given annotator has only labelled onset and offset segments instead of specific AU intensities for all the frames. The number of extracted segments for each AU is indicated in Table \ref{tab:AUPerformance}. To compute the facial descriptors at each frame, we use the same procedure described in Sec. \ref{sec:unbc_dataset}. 

\begin{figure}[t]
     \centering
     \includegraphics[width=0.43\textwidth]{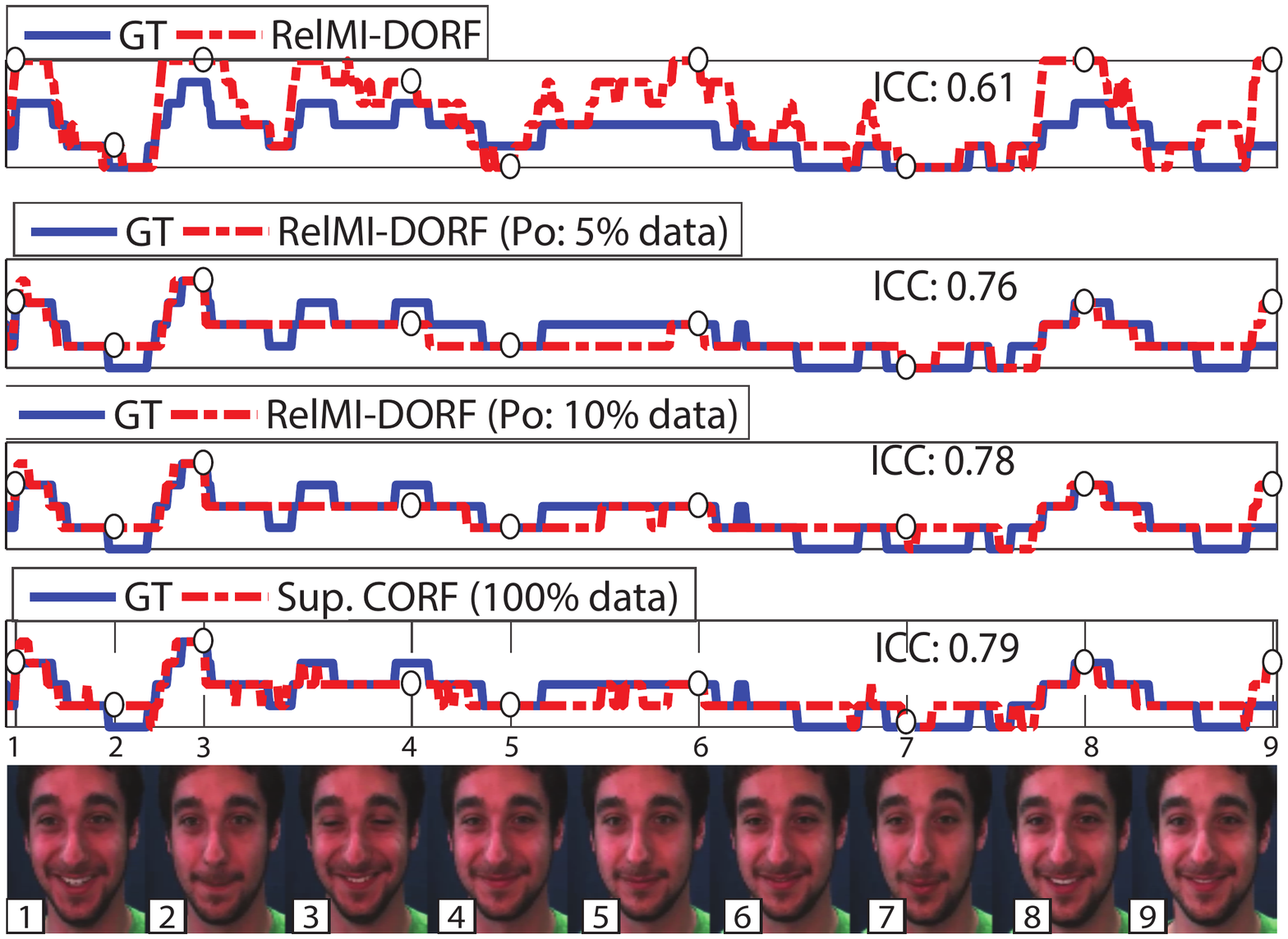}
     \caption{{Visualization of AU12 (Lip-Corner puller) intensity predictions in a subsequence of the DISFA dataset. From top to bottom: RelMI-DORF without using instance labels and with 5$\%$ and 10$\%$ of annotated frames. Supervised CORF using all the frame labels during training. Intensity estimation for RelMI-DORF tends to be more accurate as more instance labels are considered during training. Using only a $10\%$ of annotated frames, RelMI-DORF achieves similar accuracy than a fully-supervised CORF.}}
     \label{fig:DISFAQual}
\end{figure}

\subsubsection{Experimental setup and results} 
Using the  segments for each AU, we evaluate the different methods using a subject-independent 5-fold cross validation. Specifically, 3 folds are used for training and 1 for testing and validation purposes. During testing, the trained models are evaluated on the original non-segmented videos. The motivation is that, in a real scenario, onset and apex segmentation is not known for testing sequences. We also consider the partially-observed setting,  where labels for $5\%$ and $10\%$ of frames are available during training (PoRelMI-DOR). Table \ref{tab:disfa_table} shows the performance obtained by the evaluated methods computed as the average for all the considered AUs. Specific results in terms of ICC for independent AUs are shown in Table \ref{tab:AUPerformance}.

\begin{table*}[ht]
\caption{Results (ICC) for independent AUs in the DISFA Database. In parentheses, number of onset and apex segments extracted}

\centering
\resizebox{0.85\textwidth}{!}{
\begin{threeparttable}
\begin{tabular}{cccccccccccccc|c}
 &   & \textbf{AU1} & \textbf{AU2} & \textbf{AU4} & \textbf{AU5} & \textbf{AU6} & \textbf{AU9} & \textbf{AU12} & \textbf{AU15} & \textbf{AU17} & \textbf{AU20} & \textbf{AU25} & \textbf{AU26} & \textbf{AVG} \\
\textbf{Setting} & \textbf{Method}  & \textbf{(342)} & \textbf{(230)} & \textbf{(572)} & \textbf{(216)} & \textbf{(364)} & \textbf{(159)} & \textbf{(642)} & \textbf{(210)} & \textbf{(575)} & \textbf{(199)} & \textbf{(800)} & \textbf{(723)} & ~ \\ \hline

\multicolumn{1}{c|}{\multirow{5}{*}{\textbf{RelMI-DOR}}} & \multicolumn{1}{c|}{\textbf{HCRF \cite{quattoni2007hidden}}} & 0.06 & 0.03 & 0.11 & 0.01 & 0.03 & 0.02 & 0.14 & 0.01 & 0.06 & 0.01 & 0.45 & \textbf{0.24} & 0.10 \\
\multicolumn{1}{c|}{} & \multicolumn{1}{c|}{\textbf{HCORF \cite{kim2010hidden}}} & 0.02 & 0.01 & 0.05 & 0.08 & 0.02 & 0.01 & 0.04 & 0.01 & 0.01 & 0.00 & 0.06 & 0.01 & 0.03 \\
\multicolumn{1}{c|}{} & \multicolumn{1}{c|}{\textbf{OSVR \cite{zhao2016facial}}} & 0.10 & 0.13 & 0.21 & 0.04 & 0.16 & 0.09 & 0.40 & \textbf{0.09} & 0.04 & 0.04 & 0.37 & 0.17 & 0.15 \\
\multicolumn{1}{c|}{} & \multicolumn{1}{c|}{\textbf{RMI-HCRF}} & 0.02 & 0.04 & 0.10 & 0.03 & 0.12 & 0.01 & 0.30 & 0.04 & -0.02 & 0.02 & 0.40 & 0.22 & 0.11 \\
\multicolumn{1}{c|}{} & \multicolumn{1}{c|}{\textbf{RMI-DORF}} & \textbf{0.34} & \textbf{0.30} & \textbf{0.27} & \textbf{0.17} & \textbf{0.30} & \textbf{0.10} & \textbf{0.60} & 0.07 & \textbf{0.08} & \textbf{0.04} & \textbf{0.70} & 0.21 & \textbf{0.26} \\ \hline
\multicolumn{1}{c|}{\multirow{7}{*}{\textbf{\begin{tabular}[c]{@{}c@{}}PoRelMI-DOR\\ (5\% of frames)\end{tabular}}}} & \multicolumn{1}{c|}{\textbf{PoCRF \cite{li2009extracting}}} & 0.24 & 0.33 & 0.18 & 0.17 & 0.40 & 0.07 & 0.71 & 0.14 & 0.13 & 0.08 & 0.85 & 0.21 & 0.29 \\
\multicolumn{1}{c|}{} & \multicolumn{1}{c|}{\textbf{PoCORF \cite{li2009extracting}*}} & 0.20 & 0.39 & 0.21 & 0.26 & 0.41 & 0.10 & 0.77 & 0.14 & \textbf{0.15} & \textbf{0.11} & 0.80 & 0.32 & 0.32 \\
\multicolumn{1}{c|}{} & \multicolumn{1}{c|}{\textbf{PoHCRF \cite{chang2009learning}}} & 0.26 & 0.35 & 0.18 & 0.17 & 0.42 & 0.08 & 0.72 & 0.10 & 0.13 & 0.08 & \textbf{0.86} & 0.23 & 0.30 \\
\multicolumn{1}{c|}{} & \multicolumn{1}{c|}{\textbf{PoHCORF \cite{chang2009learning}}*} & 0.24 & 0.34 & 0.25 & \textbf{0.30} & 0.40 & 0.10 & \textbf{0.78} & 0.15 & \textbf{0.15} & \textbf{0.11} & 0.81 & 0.35 & 0.33 \\
\multicolumn{1}{c|}{} & \multicolumn{1}{c|}{\textbf{OSVR \cite{zhao2016facial}}} & 0.15 & 0.20 & \textbf{0.30} & 0.16 & 0.34 & 0.11 & 0.73 & 0.16 & 0.09 & \textbf{0.09} & 0.78 & \textbf{0.37} & 0.29 \\
\multicolumn{1}{c|}{} & \multicolumn{1}{c|}{\textbf{RMI-HCRF}} & 0.12 & 0.39 & 0.04 & 0.18 & \textbf{0.51} & 0.10 & 0.24 & 0.17 & 0.06 & 0.09 & 0.25 & 0.14 & 0.19 \\
\multicolumn{1}{c|}{} & \multicolumn{1}{c|}{\textbf{RMI-DORF}} & \textbf{0.38} & \textbf{0.47} & 0.28 & 0.29 & 0.44 & \textbf{0.11} & \textbf{0.78} & \textbf{0.18} & \textbf{0.15} & \textbf{0.11} & 0.78 & 0.35 & \textbf{0.36} \\ \hline
\multicolumn{1}{c|}{\multirow{7}{*}{\textbf{\begin{tabular}[c]{@{}c@{}}PoRelMI-DOR\\ (10\% of frames)\end{tabular}}}} & \multicolumn{1}{c|}{\textbf{PoCRF \cite{li2009extracting}}} & 0.27 & 0.44 & 0.21 & 0.19 & 0.46 & 0.06 & 0.72 & \textbf{0.22} & 0.16 & 0.07 & \textbf{0.84} & 0.23 & 0.32 \\
\multicolumn{1}{c|}{} & \multicolumn{1}{c|}{\textbf{PoCORF \cite{li2009extracting}*}} & 0.26 & 0.45 & 0.28 & 0.32 & 0.39 & 0.11 & 0.76 & 0.17 & 0.09 & 0.09 & 0.78 & 0.31 & 0.33 \\
\multicolumn{1}{c|}{} & \multicolumn{1}{c|}{\textbf{PoHCRF \cite{chang2009learning}}} & 0.36 & 0.46 & 0.20 & 0.24 & 0.40 & 0.08 & 0.73 & 0.26 & 0.12 & 0.08 & 0.84 & 0.29 & 0.34 \\
\multicolumn{1}{c|}{} & \multicolumn{1}{c|}{\textbf{PoHCORF \cite{chang2009learning}}*} & 0.25 & 0.44 & 0.26 & 0.35 & 0.42 & 0.11 & 0.77 & 0.20 & 0.16 & 0.09 & 0.78 & 0.32 & 0.35 \\
\multicolumn{1}{c|}{} & \multicolumn{1}{c|}{\textbf{OSVR \cite{zhao2016facial}}} & 0.15 & 0.22 & 0.29 & 0.17 & 0.34 & \textbf{0.13} & 0.74 & 0.17 & 0.10 & 0.09 & 0.77 & \textbf{0.37} & 0.29 \\
\multicolumn{1}{c|}{} & \multicolumn{1}{c|}{\textbf{RMI-HCRF}} & 0.28 & 0.44 & 0.24 & 0.21 & \textbf{0.49} & 0.08 & 0.71 & 0.20 & 0.14 & 0.12 & 0.72 & 0.22 & 0.32 \\
\multicolumn{1}{c|}{} & \multicolumn{1}{c|}{\textbf{RMI-DORF}} & \textbf{0.39} & \textbf{0.50} & \textbf{0.29} & \textbf{0.39} & 0.44 & 0.12 & \textbf{0.78} & 0.21 & \textbf{0.17} & \textbf{0.11} & 0.81 & 0.32 & \textbf{0.38} \\ \hline
\multicolumn{1}{c|}{\multirow{2}{*}{\textbf{\begin{tabular}[c]{@{}c@{}}Supervised\\ DOR\end{tabular}}}} & \multicolumn{1}{c|}{\textbf{CRF \cite{lafferty2001conditional}}} & 0.33 & 0.44 & 0.26 & 0.33 & \textbf{0.51} & 0.08 & 0.74 & \textbf{0.24} & \textbf{0.14} & \textbf{0.11} & \textbf{0.84} & 0.24 & 0.35 \\
\multicolumn{1}{c|}{} & \multicolumn{1}{c|}{\textbf{CORF \cite{kim2010structured}}} & \textbf{0.40} & \textbf{0.47} & \textbf{0.28} & \textbf{0.35} & 0.45 & \textbf{0.11} & \textbf{0.78} & 0.20 & \textbf{0.14} & 0.09 & 0.81 & \textbf{0.32} & \textbf{0.37}
\end{tabular}
\begin{tablenotes}
(*)Indicates a nominal method that we have extended to the ordinal case.
\end{tablenotes}
\end{threeparttable}
}
\label{tab:AUPerformance}
\end{table*}

\subsubsection{Discussion} 
When instance labels are not used during training (Relative MI-DOR setting), we can observe that HCRF and HCORF obtain poor results compared to OSVR and RelMI-DORF. This can be explained because the former methods explicitly model the increasing/decreasing intensity constraints provided by sequence weak-labels. Moreover, the low results obtained by RelMI-DRF compared to RelMI-DORF suggest that modelling intensities as nominal variables is suboptimal in this scenario.  Also note that OSVR obtains worse results in terms of ICC and MAE compared to RelMI-DORF. Given that performances in terms of CORR are more similar, it shows the limitation of OSVR to predict the actual scale of instance ordinal labels. Considering the results for independent AUs, we observe that RelMI-DORF achieves the best performance for most cases. Note however, that results  for some particular AUs (9,15,17, 20) is low for all the methods. We attribute this to the fact that, the activation of these AUs is typically more subtle and high-intensity levels are scarce.

By looking into the results in the partially-observed setting, we can derive the following conclusions. Firstly, all the methods improve their average performance as the percentage of instance labels increases. However, this improvement is more significant for ICC and MAE. This shows that, when instance labels are not available during training, the tendency of intensity levels can be captured. However, accurate predictions of particular ordinal labels requires the additional information provided by frame-by-frame annotations. To illustrate this, in Fig. \ref{fig:DISFAQual} we show AU12 predictions attained by RelMI-DORF using different percentages of annotated frames. Secondly, note that approaches modelling the ordinal structure of labels usually achieves better performance than nominal methods in terms of ICC and CORR. In contrast, CRF and HCRF obtain lower MAE than CORF and HCORF. This can be explained because the majority of sequence frames  has AU intensity level of 0 (neutral). As a consequence, CRF and HCRF tends to assign most of the frames to this level, thus minimizing the absolute error. In contrast, ordinal methods are more robust to imbalanced intensity levels and capture better changes in AU intensities. Finally, note that the proposed RelMI-DORF method obtain the best average performance considering $5\%$ and $10\%$ of annotated frames. Regarding specific AUs, RelMI-DORF obtain better results for most cases and competitive performance against the best method otherwise. Finally, note that RelMI-DORF performance with $10\%$ of annotated frames is comparable to the achieved by the fully-supervised approaches CRF and CORF. Specifically, only supervised CRF outperforms RelMI-DORF in terms of average MAE. The slightly worse results of supervised CORF compared with RelMI-DORF suggest that considering intensity annotations for all the frames may cause overfitting and decrease performance on unseen test sequences. This can be seen more clearly by looking at the results of independent AUs, where RelMI-DORF obtain slightly better performance than fully-supervised CORF in some cases. In conclusion, the presented results support our hypothesis that it is possible to use the proposed RelMI-DORF model in order to reduce the annotation effort required for AU intensity estimation.


\section{Conclusions and discussion}
In this work, we have presented MI-DORF for the novel task of Multi-Instance Dynamic-Ordinal Regression. To the best of our knowledge, this is the first MIL approach that imposes an ordinal structure on instance labels, and also attains dynamic modeling within bag instances. By considering different weak-relations between instance and bag labels, we have developed two variants of this framework: RelMI-DORF and MaxMI-DORF. Moreover, we have extended the proposed framework for Partially-Observed MI-DOR problems, where a subset of instance labels are also available during training. 
Although the presented MI-DORF framework has many potential applications in multiple domains, our results in the context of weakly-supervised facial behavior analysis are relevant in several aspects. In the MI-DOR setting, where no instance-level annotations are available during training, we showed that the proposed method can learn underlying variables that are significantly correlated with the ground-truth instance labels. Even though our results in this setting are lower than fully-supervised approaches, our method provides a good trade-off between the annotation effort and the accuracy of intensity predictions. While we do not claim to replace the AU/Pain annotation process using only weak-labels at sequence-level, this setting may be preferable in some applications. For example, when the focus is on capturing the variation in target facial behaviour rather than obtaining highly accurate frame labels (e.g., for monitoring changes in patient's pain intensity levels), our approach has clear advantages over the fully supervised methods which require a time-consuming annotation process. On the other hand, the competitive results of Partially-Observed MI-DORF compared to the evaluated fully-supervised approaches, indicate that annotation effort can be highly-reduced when combined with weak-information. 

{It is also worth mentioning recent works on Deep Learning for Action Unit detection \cite{tHoser2016deep} and Intensity Estimation \cite{zhou2016recurrent,walecki2017_deep_copula}. Although these models have a high modelling power, the reported results have not shown significant improvements compared to traditional shallow methods using hand-crafted features. For example, the recently proposed Copula Convolutional Neural Network (CNN) \cite{walecki2017_deep_copula} for AU Intensity Estimation is highly-related to our approach, because it combines a CNN with a probabilistic graphical model similar to the one employed in MI-DORF. Even though the Copula CNN requires intensity labels for all the frames during training, the reported results on the DISFA dataset are comparable to those achieved by our method. Specifically, MI-DORF trained with only a 10$\%$ of annotated frames obtains better average performance in terms of Mean Average Error (0.48 vs. 0.61) whereas it is outperformed in terms of ICC (0.45 vs. 0.38) (Table \ref{tab:disfa_table}). Although these results are not directly comparable because of different experimental settings, they indicate that our method trained with labels at sequence-level and a small portion of labelled frames can still show competitive performance. It is known that Supervised Deep Learning models require a large number of samples to be effectively trained \cite{han2016incremental}. Thus, this still limits their application to Facial Behavior Analysis, where the annotation process is laborious and labelled data is scarce. Posing the facial expression intensity estimation as a weakly-supervised learning problem would provide an opportunity to replace the limited-size datasets currently used in the field, by large-scale  not-fully labelled databases. Therefore, coupling Deep models with the proposed framework is a natural step forward and will be the focus of our future research. This would provide a principled way to train these powerful models by taking advantage of data-driven MIL assumptions and a vast amount of weakly-annotated data.}

\bibliographystyle{IEEEtran}

\bibliography{biblio.bib}

\vspace{-10mm}

\end{document}